\documentclass[sigconf]{acmart}

\AtBeginDocument{%
  \providecommand\BibTeX{{%
    \normalfont B\kern-0.5em{\scshape i\kern-0.25em b}\kern-0.8em\TeX}}}

\copyrightyear{2020} 
\acmYear{2020} 
\setcopyright{acmcopyright}\acmConference[KDD '20]{Proceedings of the 26th ACM SIGKDD Conference on Knowledge Discovery and Data Mining}{August 23--27, 2020}{Virtual Event, CA, USA}
\acmBooktitle{Proceedings of the 26th ACM SIGKDD Conference on Knowledge Discovery and Data Mining (KDD '20), August 23--27, 2020, Virtual Event, CA, USA}
\acmPrice{15.00}
\acmDOI{10.1145/3394486.3403192}
\acmISBN{978-1-4503-7998-4/20/08}

\usepackage{thmtools,thm-restate}
\usepackage[algo2e,ruled]{algorithm2e}

\usepackage{enumitem}
\setitemize{noitemsep,topsep=0pt,parsep=0pt,partopsep=0pt}

\usepackage{amsmath,bm,mathtools,amsfonts}
\usepackage{amsthm}

\newtheorem{remark}{Remark}
\newtheorem{assumption}{Assumption}
\newcommand{\cE}{\mathbb{E}}

\newcommand{\cO}{\mathcal{O}}

\newcommand{\sprod}{\mathop{\textstyle\prod}}
\usepackage{multirow}
\begin{document}
\fancyhead{}
\title{Minimal Variance Sampling with Provable Guarantees for Fast Training of Graph Neural Networks}

\author{Weilin Cong}
\affiliation{\institution{The Pennsylvania State University}}
\email{wxc272@psu.edu}

\author{Rana Forsati}
\affiliation{\institution{Microsoft Bing}}
\email{raforsat@microsoft.com}

\author{Mahmut Kandemir}
\affiliation{\institution{The Pennsylvania State University}}
\email{mtk2@psu.edu}

\author{Mehrdad Mahdavi}
\affiliation{\institution{The Pennsylvania State University}}
\email{mzm616@psu.edu}

\begin{abstract}
Sampling methods (e.g., node-wise, layer-wise, or subgraph) has become an indispensable strategy to speed up training large-scale Graph Neural Networks (GNNs). However, existing sampling methods are mostly based on the graph structural information and ignore the dynamicity of optimization, which leads to high variance in estimating the stochastic gradients. The high variance issue can be very pronounced in extremely large graphs, where it results in slow convergence and poor generalization.  In this paper, we theoretically analyze the variance of sampling methods and show that, due to the composite structure of empirical risk, the variance of any sampling method can be decomposed into  \textit{embedding approximation variance} in the forward stage and \textit{stochastic gradient variance} in the backward stage that necessities mitigating both types of variance  to obtain faster convergence rate. We propose a decoupled  variance reduction strategy that employs  (approximate) gradient information to adaptively sample nodes with minimal variance, and explicitly reduces the variance introduced by embedding approximation. We show theoretically and empirically that the proposed method, even with smaller mini-batch sizes, enjoys  a faster convergence rate and entails a better generalization compared to the existing  methods. 
Code is public available at~\href{https://github.com/CongWeilin/mvs_gcn}{here}.
\footnote{Please notice that we fixed a typo of our objective function defined in Eq.~\ref{equation:sgcn} on $09/05/2021$.}
\end{abstract}


\maketitle

\section{Introduction}\label{section:intro}
Graph Neural Networks (GNNs) are powerful models for learning representation of nodes  and have achieved great success in dealing with graph-related applications using data that contains rich relational information among objects, including social network prediction~\cite{kipf2016semi,hamilton2017inductive,wang2019mcne,deng2019learning,qiu2018deepinf}, traffic prediction~\cite{cui2019traffic,rahimi2018semi,li2019predicting,kumar2019predicting}, knowledge graphs~\cite{wang2019knowledge,wang2019kgat,park2019estimating}, drug reaction~\cite{do2019graph,duvenaud2015convolutional} and recommendation system~\cite{berg2017graph,ying2018graph}.

Despite the potential of GNNs, training GNNs on large-scale graphs remains a big challenge, mainly due to the inter-dependency of nodes in a graph. In particular, in GNNs, the representation (embedding) of a node is obtained by gathering the embeddings of its neighbors from the previous layers. Unlike other neural networks that the final output and gradient can be perfectly decomposed over individual data samples, in GNNs, the embedding of a given node depends recursively on all its neighbor's embedding, and such dependency grows exponentially with respect to the number of layers, a phenomenon known as \emph{neighbor explosion}, which prevents their application to large-scale graphs.  To alleviate the computational burden of training GNNs,  mini-batch sampling methods, 
including node-wise sampling~\cite{hamilton2017inductive,ying2018graph}, layer-wise sampling~\cite{zou2019layer,chen2018fastgcn,li2018adaptive}, and subgraph sampling~\cite{zeng2019graphsaint,chiang2019cluster}, have been proposed that only aggregate the embeddings of a sampled subset of neighbors of each node in the mini-batch at every layer.

Although empirical results show that the aforementioned sampling methods can scale GNN training to a large graph, these methods incur a high variance that deteriorates the convergence rate and leads to a poor generalization. To reduce the variance of sampling methods, we could either increase the mini-batch size per layer or employ adaptive sampling methods (gradient information or representations) to reduce the variance. The computation and memory requirements are two key barriers to increase the number of sampled nodes per layer in a sampled mini-batch. 

In importance sampling or adaptive sampling methods, the key idea is to utilize the gradient information which changes
during optimization to sample training examples (e.g., nodes in GNNs) to effectively reduce the variance in \textit{unbiased} stochastic gradients. Recently, different adaptive sampling methods  are proposed in the literature to speed up vanilla Stochastic Gradient Descent (SGD), 
e.g., 
importance sampling~\cite{zheng2014randomized},
adaptive importance sampling~
\cite{papa2015adaptive,csiba2015stochastic},
gradient-based sampling~\cite{papa2015adaptive,zhao2015stochastic,zhu2016gradient},
safe adaptive sampling~\cite{stich2017safe},  
bandit sampling~\cite{salehi2018coordinate}, and determinantal point processes based sampling~\cite{zhang2017determinantal}-- to name a few. Although adaptive sampling methods have achieved promising results for training neural networks via SGD, the generalization of these methods to GNNs is not straightforward. As we will elaborate later, the key difficulty is the multi-level composite structure of the training loss in GNNs, where unlike standard empirical risk minimization, any sampling idea to overcome neighbor explosion introduces a significant bias due to estimating embedding of nodes in different layers, which makes it difficult to accurately estimate the optimal sampling distribution.

\begin{figure}
    \centering 
    \includegraphics[width=0.45\textwidth]{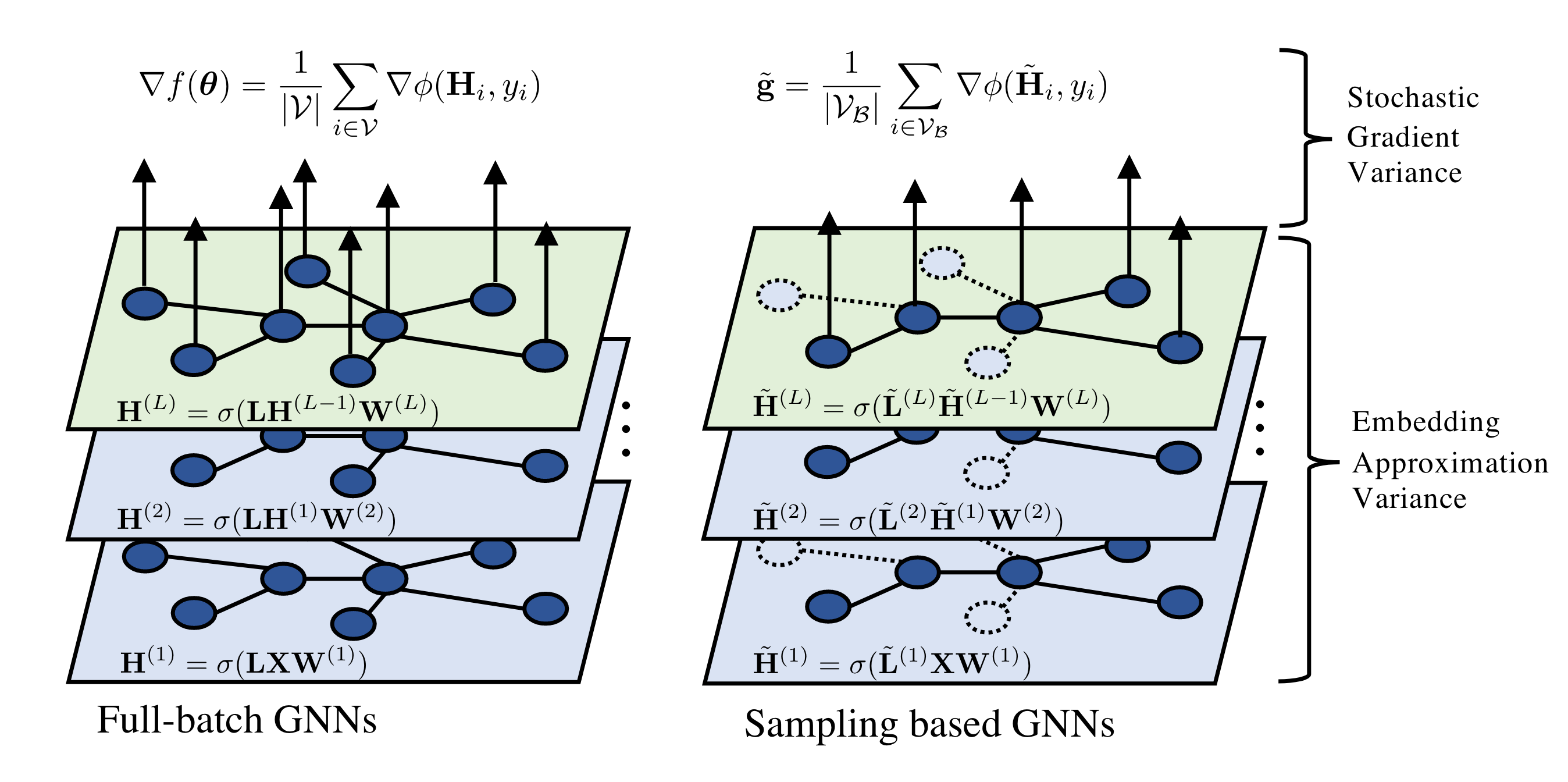}
    \caption{Comparing full-batch GNNs versus sampling based GNNs.  The sampling based GNNs incurs two types of variance: embedding approximation variance and stochastic gradient variance.}
    \label{fig:2_type_vars}
    \vspace{-0.4cm}
\end{figure}

The overarching goal of this paper is to develop a novel \textbf{decoupled variance reduction} schema that significantly reduces the variance of sampling based methods in training GNNs, and enjoys the beneficial properties of adaptive importance sampling methods in standard SGD. The motivation behind the proposed schema stems from our theoretical analysis of the variance of the sampled nodes. Specifically, we show that due to the composite structure of the training objective, the stochastic gradient is a biased estimation of the full-batch gradient that can be decomposed into two types of variance: embedding approximation variance and stochastic gradient variance. As shown in Figure~\ref{fig:2_type_vars}, embedding approximation variance exists because a subset of neighbors are sampled in each layer to estimate the exact node embedding matrix, while stochastic gradient variance exists because a mini-batch is used to estimate the full-batch gradient (similar to vanilla SGD). Besides, the bias of the stochastic gradient is proportional to the embedding approximation variance, and the stochastic gradient becomes unbiased as embedding approximation variance reduces to zero. 

The proposed minimal variance sampling schema, dubbed as \texttt{MVS-GNN},   employs the dynamic information during optimization to sample nodes and composes of two key ingredients: (i) explicit embedding variance reduction by utilizing the history of embeddings of nodes, (ii) gradient-based minimal variance sampling by utilizing the (approximate) norm of the gradient of nodes and solving an optimization problem. The proposed schema can be efficiently computed and is always better than uniform sampling or static importance sampling, as we demonstrate theoretically.  We empirically compare \texttt{MVS-GNN} through various experimental results on different large-scale real graph datasets and different sampling methods, where \texttt{MVS-GNN} enjoys a faster convergence speed by significantly reducing the variance of stochastic gradients even when  significantly smaller mini-batches are employed. Our empirical studies also corroborates the efficiency of proposed algorithm to achieve better accuracy compared to competitive methods.

\noindent\textbf{Organization.~}The remainder of this paper is organized as follows.
In Section~\ref{section:related}, we review related literature on different sampling methods to train  GNNs. In Section~\ref{section:warmup}, we provide the analysis of variance of the structural based sampling methods. In Section~\ref{section:method}, we propose a decoupled variance reduction algorithm and analyze its variance. Finally, we empirically verify the proposed schema  in Section~\ref{section:exps} and conclude the paper in Section~\ref{section:conclusion}.

\section{Additional Related Work}\label{section:related}

A key idea to alleviate the \emph{neighbor explosion} issue in GNNs is to sample a mini-batch of nodes and a subset of their neighbors at each layer to compute the stochastic gradient at each iteration of SGD.  Recently, different sampling strategies with the aim of reducing variance are proposed. For instance, node-wise sampling is utilized in \texttt{GraphSage}~\cite{hamilton2017inductive}  to restrict the computation complexity by uniformly sampling a subset of nodes from the previous layer's neighbors. However, the variance of nodes' embedding might be significantly large if the number of sampled neighbors is small. \texttt{VRGCN}~\cite{chen2017stochastic} further restricted the neighborhood size by requiring only two support nodes in the previous layer, and used the historical activation of the previous layer to reduce variance. Though successfully achieved comparable convergence as \texttt{GraphSage}, the computation complexity is high as additional graph convolution operations are performed on historical activation to reduce variance.  More importantly, node-wise sampling methods require sample nodes recursively for each node and each layer, which results in a significant large sample complexity.

Instead of performing node-wise sampling, layer-wise sampling methods, such as \texttt{FastGCN}~\cite{chen2018fastgcn}, independently sample nodes using importance sampling, which results in a constant number of nodes with low variance in all layers. However, since the sampling operation is conduced independently at each layer, it requires a large sample size to guarantee the connectivity between the sampled nodes at different layers.
LADIES~\cite{zou2019layer} further improve the sample density and reduce the sample size by restricting the candidate nodes in the union of the neighborhoods of the sampled nodes in the upper layer. However, they need to track the neighbors of nodes in the previous layer and calculate a new importance sampling distribution for each layer.

Another direction of research uses  subgraph sampling. For instance, \texttt{ClusterGCN}~\cite{chiang2019cluster} proposed to first partition graph into densely connected clusters during pre-processing, then construct mini-batches by randomly selecting subset of clusters  during training. However, its performance is significantly sensitive to the cluster size, and performing graph partition of a large graph is time-consuming. \texttt{GraphSaint}~\cite{zeng2019graphsaint} proposed to construct mini-batches by importance sampling, and apply normalization techniques to eliminate bias and reduce variance. However, since the sampling operation is conducted independently for each node, it cannot guarantee the connectivity between nodes in the sampled subgraph, which incurs a large variance due to the approximate embedding.

\section{Problem Statement}\label{section:warmup}
In this section, we formally define the problem and present a mathematical derivation of the variance of sampling strategies.

\subsection{Problem definition}
Suppose we are given a graph $\mathcal{G}(\mathcal{V},\mathcal{E})$ of $N=|\mathcal{V}|$ nodes and $|\mathcal{E}|$ edges as input, where each node is associated with a feature vector and label $(\boldsymbol{x}_i, y_i)$. Let $\mathbf{X}=[\boldsymbol{x}_1,\ldots,\boldsymbol{x}_N]$ and $\boldsymbol{y}=[y_1,\ldots,y_N]$ denote the feature matrix and labels for all $N$ nodes, respectively.
Given a $L$-layer GNN, the $\ell$th graph convolution layer is defined as $\mathbf{H}^{(\ell)} = \sigma(\mathbf{L} \mathbf{H}^{(\ell-1)} \mathbf{W}^{(\ell)}) \in \mathbb{R}^{N \times F}$,
where $\mathbf{L}$ is the normalized Laplacian matrix, $F$ is embedding dimension which we assume is the same for all layers for ease of exposition, and  $\sigma(\cdot)$ is the activation function (e.g., ReLU). Letting $\mathbf{A} \in \{0 , 1\}^{N \times N}$  and $\mathbf{D}$  be the adjacency matrix  and diagonal degree matrix associated with  $\mathcal{G}$,  the  normalized Laplacian matrix $\mathbf{L}$ is calculated as $\mathbf{L} = \mathbf{D}^{-1/2}\mathbf{A}\mathbf{D}^{-1/2}$ or $\mathbf{L} = \mathbf{D}^{-1}\mathbf{A}$. To illustrate the key ideas we focus on  the semi-supervised node classification problem, where the goal is to learn a set of per-layer weight matrices $\boldsymbol{\theta} = \{\mathbf{W}^{(1)},\ldots,\mathbf{W}^{(L)}\}$ by minimizing the empirical loss over all nodes 
\begin{equation}
    \mathcal{L}(\boldsymbol{\theta}) = \frac{1}{N} \sum_{i\in\mathcal{V}} \phi(\mathbf{H}^{(L)}_i, y_i),
    \label{eqn:main:objective}
\end{equation}
where $\phi(\cdot)$ stands for the loss function (e.g., cross entropy loss) and $\mathbf{H}^{(L)}_i$ is the node embedding of the $i$th node at the final layer computed by 
\begin{equation*}
    \mathbf{H}^{(L)} = \sigma\Big(\mathbf{L} \sigma\big(\ldots \underbrace{\sigma(\mathbf{L} \mathbf{X} \mathbf{W}^{(1)})}_{\mathbf{H}^{(1)}} \ldots\big) \mathbf{W}^{(L)}\Big).
\end{equation*}
with $\mathbf{H}^{(0)} = \mathbf{X}$ is set to be the input for the first layer. To efficiently solve the optimization problem in Eq.~\ref{eqn:main:objective} using mini-batch SGD, in the standard sampling based  methods, instead of computing the full-gradient, we only calculate an unbiased gradient based on a mini-batch $\mathcal{V}_\mathcal{B}$ of nodes with size  $B$ to update the model,
\begin{equation}
    {\mathbf{g}} = \frac{1}{B} \sum_{i\in\mathcal{V}_\mathcal{B}} \nabla \phi({\mathbf{H}}^{(L)}_i, y_i).
    \label{eqn:minibatch}
\end{equation}
However, computing the gradient in Eq.~\ref{eqn:minibatch} requires the embedding of all adjacent neighbors in the previous layers  which exponentially grows by the number of layers. A remedy is to sample a subset of nodes  at each layer to construct a sparser Laplacian matrix 
from $\mathbf{L}$ to estimate the node embedding matrices $\tilde{\mathbf{H}}^{(\ell)}$ for $\ell=1,2,\ldots, L$,  that results in a much lower computational and memory complexities for training.  


In node-wise sampling (e.g., \texttt{GraphSage}, \texttt{VRGCN}), the main idea is to first sample all the nodes needed for the computation using neighbor sampling (NS), and then update the parameters. Specifically, for each node in the $\ell$th layer, NS randomly samples $s$ of its neighbors at   $(\ell-1)$th layer and formulate $\tilde{\mathbf{L}}^{(\ell)}$ by 
\begin{equation}\label{eq:node_wise}
    \tilde{L}_{i,j}^{(\ell)} = \begin{cases}
 \frac{|\mathcal{N}(i)|}{s} \times L_{i,j}, &\text{ if } j \in \widehat{\mathcal{N}}^{(\ell)}(i) \\
 0, &\text{ otherwise }
\end{cases},
\end{equation}
where $\mathcal{N}(i)$ is  full set  of the $i$th node neighbor, $\widehat{\mathcal{N}}^{(\ell)}(i)$ is the sampled neighbors of node $i$ for $\ell$th GNN layer.

In layer-wise sampling (e.g., \texttt{FastGCN}, \texttt{LADIES}), the main idea is to control the size of sampled neighborhoods in each layer.  For the $\ell$th layer, layer-wise sampling methods sample a set of nodes $\mathcal{V}_\ell\subseteq \mathcal{V}$ of size $s$ under a distribution $\boldsymbol{p} \in  \mathbb{R}_{+}^{|\mathcal{V}|}, \sum_{i}p_i = 1$ to approximate the Laplacian by
\begin{equation}\label{eq:layer-wise}
    \tilde{L}_{i,j}^{(\ell)} = \begin{cases}
 \frac{1}{s \times p_j}\times L_{i,j}, &\text{ if } j \in \mathcal{V}_\ell \\
 0, &\text{ otherwise }
\end{cases}
\end{equation}
Subgraph sampling (e.g., \texttt{GraphSaint}, \texttt{ClusterGCN}) is similar to layer-wise sampling by restricting $\tilde{\mathbf{L}}^{(1)}=\tilde{\mathbf{L}}^{(2)}=\ldots=\tilde{\mathbf{L}}^{(L)}$.

\subsection{Variance analysis}
While being computationally appealing, the key issue that sampling methods suffer from is the additional bias introduced to the stochastic gradients due to the approximation of node embeddings at different layers.  
To concretely understand this bias, let us formulate a $L$-layer sampling based GNN as a multi-level composite stochastic optimization problem of the following form 
\begin{equation}\label{equation:sgcn}
    \min f(\boldsymbol{\theta}) := 
    \mathbb{E}_{\omega_L}\Big[ f_{\omega_L}^{(L)} \Big( 
    \mathbb{E}_{\omega_{L-1}}\big[ f_{\omega_{L-1}}^{(L-1)} \big( 
    \ldots 
    \mathbb{E}_{\omega_1} [ f_{\omega_1}^{(1)} ( \boldsymbol{\theta} )]
    \ldots
    \big) \big]
    \Big) \Big],
\end{equation}
where the random variables $\omega_\ell$ capture the stochasticity due to sampling of nodes   at the $\ell$th layer, i.e.,  the deterministic function at $\ell$th layer  $f^{(\ell)}(\boldsymbol{\theta}) := \sigma(\mathbf{L} \mathbf{H}^{(\ell-1)} \mathbf{W}^{(\ell)})$ and  its stochastic variant   $f_{\omega_\ell}^{(\ell)}(\boldsymbol{\theta}) := \sigma(\tilde{\mathbf{L}}^{(\ell)} \tilde{\mathbf{H}}^{(\ell-1)} \mathbf{W}^{(\ell)})$ induced by $\omega_\ell$. We denote the deterministic composite function at $\ell$th layer by $F^{(\ell)}(\cdot) := f^{(\ell)}\circ f^{(\ell-1)} \circ \ldots \circ f^{(1)}(\cdot)$.
By the chain rule, the full gradient can be computed   as $\nabla f(\boldsymbol{\theta}) = \nabla f^{(1)}(\boldsymbol{\theta}) \cdot \nabla f^{(2)}(F^{(1)}(\boldsymbol{\theta})) \ldots \nabla f^{(L)}(F^{(L-1)}(\boldsymbol{\theta}))$.
%
For a given sample path $(\omega_1,\ldots,\omega_L)$, one may formulate an unbiased estimate of $\nabla f(\boldsymbol{\theta})$ as $\mathbf{g} = \nabla f_{\omega_1}^{(1)}(\boldsymbol{\theta}) \cdot \nabla f_{\omega_2}^{(2)}(F^{(1)}(\boldsymbol{\theta})) \ldots \nabla f_{\omega_L}^{(L)}(F^{(L-1)}(\boldsymbol{\theta}))$,
which cannot be calculated because $F^{(\ell)}(\boldsymbol{\theta}) = f^{(\ell)}\circ f^{(\ell-1)}\circ \ldots \circ f^{(1)}(\boldsymbol{\theta})$ for $\ell\geq2$ are unfortunately not known.
In other words, the stochastic gradient $\tilde{\mathbf{g}}$ is a biased estimation of $\nabla f(\boldsymbol{\theta})$, where $\tilde{\mathbf{g}} := \nabla f^{(1)}_{\omega_1}(\boldsymbol{\theta}) \nabla f^{(2)}_{\omega_2}(f_{\omega_1}^{(1)}(\boldsymbol{\theta})) \ldots \nabla f^{(L)}_{\omega_L}(f_{\omega_{L-1}}^{(L-1)}\circ \ldots \circ f_{\omega_1}^{(1)}(\boldsymbol{\theta}))$.
We note that this is in contrast to the standard SGD where the gradient can be decomposed over training examples; thereby, the average gradient computed at a mini-batch is an unbiased estimator of full gradient.
To outline the role of bias and variance in the stochastic gradients of training GNNs, we note that in vanilla SGD for empirical risk minimization, we assume the variance of the unbiased stochastic gradients $\mathbf{g}$ are bounded, i.e., $  \cE[\|{\mathbf{g}} - \nabla f(\boldsymbol{\theta})\|^2]$, but in GNNs due to sampling at inner layers, this no longer holds. In fact, the noise of stochastic gradient estimator $\tilde{\mathbf{g}}$,  can be decomposed as
\begin{equation*}
    \cE[\|\tilde{\mathbf{g}} - \nabla f(\boldsymbol{\theta})\|^2] = \underset{\text{bias} ~(\mathbb{V})}{\cE[\|\tilde{\mathbf{g}} - \mathbf{g}\|^2]} + \underset{\text{variance}~(\mathbb{G})}{\cE[\|\mathbf{g} - \nabla f(\boldsymbol{\theta})\|^2]},
\end{equation*}
where bias is due to the inner layers embedding approximation in forward pass, and the variance corresponds to the standard variance due to mini-batch sampling. 
We make the following standard assumption on the Lipschitz continuity of functions $f^{(\ell)}(\cdot)$.
\begin{assumption} \label{assumption:sgcn_lip_smooth}
For each $\ell=1,\ldots,L$ and each realization of $\omega_\ell$, the mapping $f_{\omega_\ell}^{(\ell)}(\cdot)$ is $\rho_\ell$-Lipschitz and its gradient $\nabla f_{\omega_\ell}^{(\ell)}(\cdot)$ is $G_\ell$-Lipschitz.
\end{assumption}

\begin{table*}[t]
\caption{Summary of function approximation variance. Here $D$ denotes the average node degree, $s$ denotes the neighbor sampling size, $N_\ell$ denotes the size of nodes sampled in $\ell$th layer, $\gamma_\ell$ denotes the upper-bound of $\|\mathbf{H}_i^{(\ell-1)}\mathbf{W}^{(\ell)}\|_2$, and $\Delta\gamma_\ell$ denotes the upper-bound of $\|(\mathbf{H}_i^{(\ell-1)}-\bar{\mathbf{H}}_i^{(\ell-1)})\mathbf{W}^{(\ell)}\|_2$ for any $i\in\mathcal{V}$. We use $\cO(\cdot)$ to hide constants that remain the same between different algorithms.}
\label{table:embedding_approximation_variance}
\centering
\begin{tabular}{l|l|l|l|l|l}
\hline
\textbf{Method}   & \texttt{GraphSage} & \texttt{VRGCN} & \texttt{LADIES} & \texttt{GraphSaint} & \texttt{MVS-GNN} \\ \hline\hline
\textbf{Variance} & $\cO(D\gamma_\ell^2/s)$ & $\cO(D\Delta\gamma_\ell^2/s)$ & $\cO(N\gamma_\ell^2/N_\ell)$  & $\cO(N^2\gamma_\ell^2/N^2_\ell)$  & $\cO(D\Delta\gamma_\ell^2)$ \\ \hline
\end{tabular}
\end{table*}


The following lemma shows that the bias of stochastic gradient  can be decomposed as a  combination of embedding approximation variance of different layers.
\begin{lemma}\label{lemma:decouple_bias}
Let $\mathbb{V}_\ell:= \cE[\|f_{\omega_\ell}^{(\ell)}(F^{(\ell-1)}(\boldsymbol{\theta})) - F^{(\ell)}(\boldsymbol{\theta})\|^2]$ be the per-layer embedding approximation variance. 
Suppose Assumption~\ref{assumption:sgcn_lip_smooth} holds. Then,  the bias of stochastic gradient $\cE[\|\mathbf{g}-\tilde{\mathbf{g}}\|^2]$ can be bounded as:
\begin{equation*}
    \cE[\|\mathbf{g}-\tilde{\mathbf{g}}\|^2] \leq L\cdot\sum_{\ell=2}^L \left(\sprod_{i=1}^{\ell-1} \rho^2_i \right)\left(\sprod_{i=\ell+1}^L \rho^2_i \right) G^2_\ell \cdot \ell \sum_{i=1}^\ell \left(\sprod_{j=i+1}^\ell \rho_j^2 \mathbb{V}_j \right). 
\end{equation*}
\end{lemma}
\begin{proof}
Proof is deferred to Appendix~\ref{section:bias_decouple_proof}.
\end{proof}
From decomposition of variance and Lemma~\ref{lemma:decouple_bias}, we conclude that any sampling method introduces two types of variance, i.e., embedding approximation variance $\mathbb{V}$   and stochastic gradient variance $\mathbb{G}$, that controls the degree of biasedness of stochastic gradients. Therefore, any  sampling strategy needs to take into account both kinds of variance to speed up the convergence. Indeed, this is one of the key hurdles in applying adaptive importance sampling methods such as  bandit sampling or gradient based importance sampling to sampling based GNN training -- originally developed for vanilla SGD, as accurate estimation of gradients is crucial to reduce the variance, which is directly affected by variance in approximating the embedding matrices at different layers. 
\begin{remark}
We emphasize that the aforementioned sampling methods are solely based on the Laplacian matrix and fail to explicitly leverage the dynamic information during training to further reduce the variance.    However, from Lemma~\ref{lemma:decouple_bias}, we know that the bias of stochastic gradient can be controlled by applying explicit variance reduction to function approximation variance $\mathbb{V}$, which motivates us developing a decoupled variance reduction algorithm to reduce the both types of variance. 
\end{remark}

\section{Adaptive minimal variance sampling}\label{section:method}

Motivated by the variance analysis in the previous section, we now present a decoupled variance reduction algorithm, \texttt{MVS-GNN}, that effectively reduces the variance in training GNNs using an adaptive importance sampling strategy by leveraging gradient and embedding information during optimization.  To sample the nodes, we propose a minimal variance sampling strategy based on the estimated norm of gradients.  To reduce the effect of embedding approximation variance in estimating the gradients, we explicitly reduce it at each layer using the history of embeddings of nodes in the previous layer.  

\subsection{Decoupled variance reduction}


The detailed steps of the proposed algorithm are summarized in Algorithm~\ref{algorithm:mvs_gnn}. To effectively reduce both types of variance, we propose an algorithm with two nested loops. In the outer-loop, at each iteration $t = 1, 2, \ldots, T$ we sample a large mini-batch $\mathcal{V}_{\mathcal{S}}$ of size $S = N \times \gamma$ uniformly at random, where $\gamma \in (0, 1]$ is the  sampling ratio,  to estimate the gradients and embeddings of nodes.  The outer-loop can be considered as a checkpoint to refresh the estimates as optimization proceeds, where   $\gamma$ controls the accuracy of estimations at the checkpoint.  Specifically, at every checkpoint, we calculate the per sample gradient norm as $\bar{\mathbf{g}} = [\bar{g}_1, \ldots,\bar{g}_{S}]$ and save it to memory for further calculation of the importance sampling distribution.

Meanwhile, we also compute the node embedding for each node in $\mathcal{V}_\mathcal{S}$. To do so, we construct $\{\tilde{\mathbf{L}}^{(\ell)}\}_{\ell=1}^L$ that only contains nodes needed for calculating  embeddings of nodes in $\mathcal{V}_\mathcal{S}$, without node-wise or layer-wise node sampling. 
Then, we calculate the node embedding $\tilde{\mathbf{H}}^{(\ell)}$ and update its history embedding $\bar{\mathbf{H}}^{(\ell)}$ as
\begin{equation}\label{eq:full_batch}
    \begin{aligned}
    \tilde{\mathbf{H}}_i^{(\ell)} = \sigma\left(\sum_{j\in\mathcal{V}} \tilde{L}^{(\ell)}_{i,j} \tilde{\mathbf{H}}_i^{(\ell-1)} \mathbf{W}^{(\ell)} \right),~\bar{\mathbf{H}}_i^{(\ell)} = \tilde{\mathbf{H}}_i^{(\ell)}.
    \end{aligned}
\end{equation}


Every iteration of outer-loop is followed by $K$ iterations  of the inner-loop, where at each iteration $k =  2, \ldots, K$, we sample a small mini-batch $\mathcal{V}_{\mathcal{B}} \subset \mathcal{V}_{\mathcal{S}}$ of size $B$, and prepare the Laplacian matrix of each layer $\{\tilde{\mathbf{L}}^{(\ell)}\}_{\ell=1}^L$ to estimate the  embeddings for nodes in $\mathcal{V}_\mathcal{B}$ and update the parameters of GNN. 
Our key idea of reducing the variance of embeddings is to use the history embeddings of nodes in the previous layer $\bar{\mathbf{H}}^{(\ell-1)}$ as a feasible approximation to estimate the node embeddings in the current layer $\tilde{\mathbf{H}}^{(\ell)}$. 
Each time when $\tilde{\mathbf{H}}_i^{(\ell)}$ is computed, we update $\bar{\mathbf{H}}_i^{(\ell)}$ with $\tilde{\mathbf{H}}_i^{(\ell)}$:
\begin{equation}\label{eq:mini_batch}
    \begin{aligned}
    \tilde{\mathbf{H}}_i^{(\ell)} &= \sigma\left(\sum_{j\in\mathcal{V}_{\ell-1}} \tilde{L}^{(\ell)}_{i,j} \tilde{\mathbf{H}}_i^{(\ell-1)} \mathbf{W}^{(\ell)} + \sum_{j\in \mathcal{V}\backslash\mathcal{V}_{\ell-1}} L_{i,j} \bar{\mathbf{H}}_i^{(\ell-1)} \mathbf{W}^{(\ell)}\right), \\
    \bar{\mathbf{H}}_i^{(\ell)} &= \tilde{\mathbf{H}}_i^{(\ell)}
    \end{aligned}
\end{equation}
The sampling of nodes in  $\mathcal{V}_{\mathcal{B}}$ is based on a novel gradient-based minimal variance strategy to compute the to optimal sampling distribution $\boldsymbol{p}$ that will be detailed later. 
After updating the parameters, we use the freshly computed gradient and embedding of nodes in $\mathcal{V}_{\mathcal{B}} $ to update the stale information.  We note that as the gradient of objective vanishes when we approach the optimal solution, we can use larger  $K$ in later steps to reduce the number of checkpoints. Besides, we only need to maintain the norm of the gradient for nodes which requires only an additional $O(N \times \gamma)$ memory which is negligible (e.g, we set  $\gamma = 0.02$ for the Yelp dataset). 

\vspace{3pt}
\noindent\textbf{Variance analysis and time complexity.}~We summarized the embedding approximation variance of different sampling based GNN training methods in Table~\ref{table:embedding_approximation_variance}. 
We provide a detailed analysis of the embedding approximation variance of \texttt{MVS-GNN} in Appendix~\ref{section:variance_analysis}. 
Comparing with \texttt{GraphSage}, \texttt{LADIES}, and \texttt{GraphSaint}, \texttt{MVS-GNN} enjoys a much smaller variance because $\|(\mathbf{H}_i^{(\ell-1)}-\bar{\mathbf{H}}_i^{(\ell-1)})\mathbf{W}^{(\ell)}\|_2$ is usually much smaller than $\|\mathbf{H}_i^{(\ell-1)}\mathbf{W}^{(\ell)}\|_2$.
On the other hand, although the embedding approximation variance of \texttt{VRGCN} is $s$ times smaller than \texttt{MVS-GNN}, since full-batch GNN are performed once a while, the staleness of $\{\mathbf{H}^{(\ell)}\}_{\ell=1}^L$ can be well controlled, which is not true in  \texttt{VRGCN}.

\begin{remark}
Since both \texttt{MVS-GNN} and \texttt{VRGCN} utilize explicit variance reduction on estimating the embedding matrix, here we emphasize the key differences:
\begin{itemize}
    \item \texttt{MVS-GNN} is one-shot sampling, i.e., it only needs to sample one time to construct a mini-batch, while \texttt{VRGCN} requires samplers to explore recursively for each layer and each node in the mini-batch. Notice that the sample complexity can be much higher than computation complexity when the graph is large.
    \item \texttt{MVS-GNN} requires a constant number of nodes at each layer, despite the fact the dependency grows exponentially with respect to the number of layers.
    \item \texttt{MVS-GNN} requires to multiply adjacency matrix with embedding matrix one time for each forward propagation, while \texttt{VRGCN} requires twice. Therefore, the computation cost of our algorithm is relatively lower, especially when the number of layers is large.
\end{itemize}
\end{remark}
\begin{algorithm2e}[t]
    \DontPrintSemicolon
    \caption{\texttt{MVS-GNN}}
    \label{algorithm:mvs_gnn}
    \textbf{input:} 
    initial point $\boldsymbol{\theta} = \{\mathbf{W}^{(1)}, \mathbf{W}^{(2)}, \ldots, \mathbf{W}^{(L)}\}$, learning rate $\eta$, 
     mini-batch size $B$, 
    importance sampling ratio $\gamma$\\
    Set $\mathbf{H}^{(0)} = \mathbf{X}$\\
    \For{$t = 1,\ldots,T$}{
        \texttt{/* Run large-batch GNN*/} \\
        Sample  $\mathcal{V}_\mathcal{S}\subseteq\mathcal{V}$ of size $S= N\times\gamma$ uniformly at random\\
        Construct $\{\tilde{\mathbf{L}}^{(\ell)}\}_{\ell=1}^L$ based on sampled nodes in  $\mathcal{V}_\mathcal{S}$ \\
        \For{$\ell=1,\ldots,L$}{
            Estimate embedding matrices using Eq.~\ref{eq:full_batch} and update history embeddings
        }
        Update parameters
            $\boldsymbol{\theta} \leftarrow \boldsymbol{\theta} - \eta \frac{1}{S} \sum_{i\in\mathcal{V}_\mathcal{S}} \frac{\nabla \phi(\tilde{\mathbf{H}}^{(L)}_i, y_i)}{p_i}$ \\
        Calculate gradient norm $\bar{\mathbf{g}} = [\bar{g}_1,\ldots,\bar{g}_S]$ where $\bar{g}_i = \|\nabla \phi(\tilde{\mathbf{H}}^{(L)}_i, y_i)\|$ \\
        \texttt{/* Run mini-batch GNN*/} \\
        \For{$k= 2,\ldots, K$}{
            Calculate the sampling distribution $\boldsymbol{p} = [p_1,\ldots,p_S]$ using Eq.~\ref{eq:calculate_p_from_g} based on $\bar{\mathbf{g}}$\\
            Sample nodes $\mathcal{V}_\mathcal{B} \subset \mathcal{V}_\mathcal{S}$ of size $B$ with probability $\boldsymbol{p}$\\
            Construct $\{\tilde{\mathbf{L}}^{(\ell)}\}_{\ell=1}^L$  for nodes in $\mathcal{V}_\mathcal{B}$ \\
            \For{$\ell=1,\ldots,L$}{
                Calculate embeddings using  Eq.~\ref{eq:mini_batch} and update history embeddings
            }
            Update parameters
            $\boldsymbol{\theta} \leftarrow \boldsymbol{\theta} - \eta \frac{1}{B} \sum_{i\in\mathcal{V}_\mathcal{B}} \frac{\nabla \phi(\tilde{\mathbf{H}}^{(L)}_i, y_i)}{p_i}$ \\
            Update $\bar{\mathbf{g}}$ for $i\in\mathcal{V}_\mathcal{B}$ using the norm of fresh gradients
        }
    }
    \textbf{output:} 
    $\boldsymbol{\theta}$
\end{algorithm2e} 
\subsection{Gradient-based minimal variance sampling}
Here we propose a minimal variance sampling strategy to reduce the stochastic gradient variance where nodes with larger gradient are chosen with higher probability than ones with smaller gradient.  To do so, recall the optimization problem for  GNN is $ f(\boldsymbol{\theta}) := \sigma\Big(\mathbf{L} \sigma\big(\ldots \sigma(\mathbf{L} \mathbf{X} \mathbf{W}^{(1)}) \ldots\big) \mathbf{W}^{(L)}\Big)$.
Let $f_i(\boldsymbol{\theta})$ as the $i$th output of $f(\boldsymbol{\theta})$.
Formally, we consider the loss function and full-gradient as $\mathcal{L}(\boldsymbol{\theta}) = \sum_{i=1}^N \phi(f_i(\boldsymbol{\theta}), y_i)$ where $\nabla \mathcal{L}(\boldsymbol{\theta}) = \sum_{i=1}^N \nabla \phi(f_i(\boldsymbol{\theta}), y_i)$.
Rather than using all samples at each steps, we sample a sequence of random variables $\{\xi_i\}_{i=1}^{N}$, where $\xi_i \sim \mathrm{Bernoulli}(p_i)$, and $\xi_i=1$ indicates that the $i$th node is sampled and should be used to calculate the stochastic gradient 
$\mathbf{g} = \sum_{i=1}^N \frac{\xi_i}{p_i} \nabla \phi(f_i(\boldsymbol{\theta}), y_i)$.
Define $\mathbb{G} = \cE[\|\mathbf{g} - \cE[\mathbf{g}]\|^2]$. For a given mini-batch size $B$, our goal is to find the best sampling probabilities $\{p_i\}_{i=1}^N$ to  minimize $\mathbb{G}$, which can be casted as the following optimization problem:
\begin{equation*}\label{equation:1}
\begin{aligned}
    \underset{p_i}{\min} & \sum_{i=1}^N \frac{1}{p_i}\|\nabla \phi(f_i(\boldsymbol{\theta}),y_i)\|^2 \\
    & \text{subject to } \sum_{i=1}^N p_i = B,~p_i\in(0,1] \text{ for all } i.
    \end{aligned}
\end{equation*}
Although this distribution can minimize the variance of the stochastic gradient, it requires the calculation of $N$ derivatives at each step, which is clearly inefficient. 
As mentioned in \cite{zhao2015stochastic,katharopoulos2018not}, a practical solution is to relax the optimization problem as follows
\begin{equation}\label{equation:1_relax}
\begin{aligned}
    \underset{p_i}{\min} & \sum_{i=1}^N \frac{\bar{g}_i^2}{p_i} \\
    & \text{subject to } \sum_{i=1}^N p_i = B,~p_i\in(0,1] \text{ for all } i,
    \end{aligned}
\end{equation}
where $\bar{g}_i \geq \|\nabla \phi(f_i(\boldsymbol{\theta}),y_i)\|$ is the upper-bound of the per-sample gradient norm as estimated in Algorithm~\ref{algorithm:mvs_gnn}. In practice, we choose to estimate $\bar{g}_i$ using the stochastic gradient of the last GNN layer.

\begin{theorem}\label{theorem:1}
 There exist a value $\mu$ such that $p_i = \min\left( 1, \frac{\bar{g}_i}{\mu} \right)$ is the solution of Eq.~\ref{equation:1_relax}.
\end{theorem}
\begin{proof}
The Lagrange function of Eq.~\ref{equation:1_relax} has form:
\begin{equation*}
    L(\alpha,\boldsymbol{\beta}, \boldsymbol{\gamma}) = \sum_{i=1}^N \frac{\bar{g}_i^2}{p_i} + \alpha\left(\sum_{i=1}^N p_i - B\right) - \sum_{i=1}^N \beta_i p_i - \sum_{i=1}^N \gamma_i(1-p_i). 
\end{equation*}

From the KKT conditions, we have
\begin{equation*}
\left\{
\begin{array}{lr}
\frac{\partial L}{\partial p_i} = -\frac{\bar{g}_i^2}{p_i^2} + \alpha - \beta_i - \gamma_i = 0 \text{ for all } i &\\ 
\beta_i p_i = 0 \text{ for all } i & \\
\gamma_i(1-p_i) = 0 \text{ for all } i &
\end{array}\right.
\end{equation*}

By examining these conditions, it is easy to conclude that optimal solution has the following properties: (a) Since every $p_i>0$, we have $\beta_i=0$ for all $i$; (b) If $\gamma_i>0$, then $p_i=1$ and $\bar{g}_i^2 > \alpha + \gamma_i > \alpha$; (c) If $\gamma_i=0$, then $p_i = \sqrt{\bar{g}_i^2/\alpha}$.

Putting all together, we know that there exist a threshold $\sqrt{\alpha}$ that divides sample into two parts:
$\{i: \bar{g}_i < \sqrt{\alpha}\}$ of size $\kappa$ with $p_i = \sqrt{\bar{g}_i^2/\alpha}$ and $\{i :\bar{g}_i > \sqrt{\alpha}\}$ of size $N-\kappa$ with $p_i=1$

Therefore, it is sufficient to find $\alpha=\alpha^\star$ such that $\sum_{i=1}^N p_i = B$. The desired value of $\alpha^\star$ can be found as a solution of  $\sum_{i=1}^N p_i = \sum_{i=1}^{\kappa} \sqrt{\frac{\bar{g}^2_i}{\alpha}} + N-\kappa = B$. We conclude the proof by setting $\mu=\sqrt{\alpha^\star}$.
\end{proof}
From Theorem \ref{theorem:1}, we know that given per-sample gradient, we can calculate a Bernoulli importance sampling distribution $\boldsymbol{p}:=\{p_i\}_{i=1}^N$ that minimize the variance.  The following lemma gives a brute-force algorithm to compute the $\mu$ which can be used to compute the optimal sampling probabilities.
\begin{lemma}\label{lemma:bf_solution}
Suppose $\bar{g}_i$ is sorted such that $0<\bar{g}_i\leq \ldots \leq \bar{g}_N$. Let $\kappa$ be the largest integer for which $B+\kappa-N \leq \bar{g}_i/(\sum_{i=1}^\kappa \bar{g}_i)$, then $\mu = (B+\kappa-N) / (\sum_{i=1}^\kappa \bar{g}_i)$, and the probabilities can be computed by
\begin{equation}\label{eq:calculate_p_from_g}
    p_i = 
    \begin{cases}
    (B+\kappa-N)\frac{\bar{g}_i}{\sum_{j=1}^\kappa \bar{g}_j} & \text{ if } i\leq \kappa \\
    1 & \text{ if } i > \kappa \\ 
    \end{cases}
\end{equation}
\end{lemma}
\begin{proof} 
The correctness of Lemma \ref{lemma:bf_solution} can be shown by plugging the result back to Theorem \ref{theorem:1}.
\end{proof}
If we assume $B \bar{g}_N \leq \sum_{i=1}^N \bar{g}_i$, then $\kappa=N$ and $p_i = B\bar{g}_i/(\sum_{i=1}^N \bar{g}_i)$. Note that this assumption can be always satisfied by uplifting the smallest $\bar{g}_i$. We now compare the variance of the proposed importance sampling method with the variance of naive uniform sampling in Lemma~\ref{lemma:us_vs_is}.

\begin{lemma}\label{lemma:us_vs_is}
Let $\boldsymbol{p}_{us} = [p_1,\ldots, p_N]$ be the uniform sampling distribution with $p_i = B/N$, and  $\boldsymbol{p}_{is} = [p_1,\ldots, p_N]$ as the minimal variance sampling  distribution with $p_i = B\bar{g}_i/(\sum_{i=1}^N \bar{g}_i)$. 
Define $\mathbb{G}(\boldsymbol{p}_{us})$ and $\mathbb{G}(\boldsymbol{p}_{is})$ as the variance of the stochastic gradient of uniform  and minimal variance sampling, respectively.
Then,  the difference between the variance of uniform sampling and importance sampling is proportion to the Euclidean distance between $\boldsymbol{p}_{us}$ and $\boldsymbol{p}_{is}$, i.e.,
\begin{equation*}
    \begin{aligned}
        \mathbb{G}(\boldsymbol{p}_{us})-\mathbb{G}(\boldsymbol{p}_{is}) 
        &= \frac{\left(\sum_{i=1}^N \bar{g}_i\right)^2}{B^3 N} \|\boldsymbol{p}_{is}-\boldsymbol{p}_{us}\|_2^2.
    \end{aligned}
\end{equation*}
\end{lemma}
\begin{proof}
Proof is deferred to Appendix~\ref{lemma:us_vs_is}.
\end{proof}
From Lemma~\ref{lemma:us_vs_is}, we observe  that the variance of importance sampling $\mathbb{G}(\boldsymbol{p}_{is})$ is smaller than the variance of uniform sampling $\mathbb{G}(\boldsymbol{p}_{us})$ if 
the optimal importance sampling distribution is different from uniform sampling distribution (the per sample gradient norm is not all the same), i.e., $\boldsymbol{p}_{is} \neq \boldsymbol{p}_{us}$ where $\boldsymbol{p}_{is}$ is defined in Eq.~\ref{eq:calculate_p_from_g}. Besides, the effect of variance reduction becomes more significant when the difference between optimal importance sampling distribution and uniform sampling distribution is large (i.e., the difference between per-sample gradient norm is large).



\subsection{Implementation challenges}\label{subsec:implemntation}
Calculating the optimal importance sampling distribution requires having access to the stochastic gradient for every example in the mini-batch. Unfortunately, existing machine learning packages, such as Tensorflow~\cite{abadi2016tensorflow} and PyTorch~\cite{paszke2019pytorch}, does not support computing gradients with respect to individual examples in a mini-batch.

A naive approach to calculate the per sample gradient of $N$ nodes is to run backward propagation $N$ times with a mini-batch size of $1$. In practice, the naive approach performs very poorly because backward propagation is most efficient when efficient matrix operation implementations can exploit the parallelism of mini-batch training.

As an alternative, we perform backward propagation only once and reuse the intermediate results of backward propagation for per sample gradient calculation.
Recall that the embedding of node $i$ at the $\ell$th GNN layer can be formulated as $\tilde{\mathbf{H}}_i^{(\ell)} = \sigma(\tilde{\mathbf{L}}_{i}^{(\ell)} \tilde{\mathbf{H}}^{(\ell-1)} \mathbf{W}^{(\ell)})$.
During the forward propagation we save the $\tilde{\mathbf{L}}_{i}^{(\ell)} \tilde{\mathbf{H}}^{(\ell-1)}$ and during backward propagation we save the $\nabla_{\tilde{\mathbf{H}}_i^{(\ell)}} \mathcal{L}(\boldsymbol{\theta})$. Then, the gradient of updating $\mathbf{W}^{(\ell)}$ is calculated as $\left(\nabla_{\tilde{\mathbf{H}}_i^{(\ell)}} \mathcal{L}(\boldsymbol{\theta})\right) \left( \tilde{\mathbf{L}}_{i}^{(\ell)} \tilde{\mathbf{H}}^{(\ell-1)}\right)$. Despite the need for additional space to store the gradient, the time it takes to obtain per sample gradient is much lower.


\section{Experiments}\label{section:exps}
In this section, we conduct experiments to evaluate \texttt{MVS-GNN} for training GNNs on large-scale node classification datasets~\footnote{The implementation of algorithms are publicly available at \href{https://github.com/CongWeilin/mvs_gcn}{\emph{here}}.}.
\begin{figure*}[t]
    \centering
    \includegraphics[width=0.95 \textwidth]{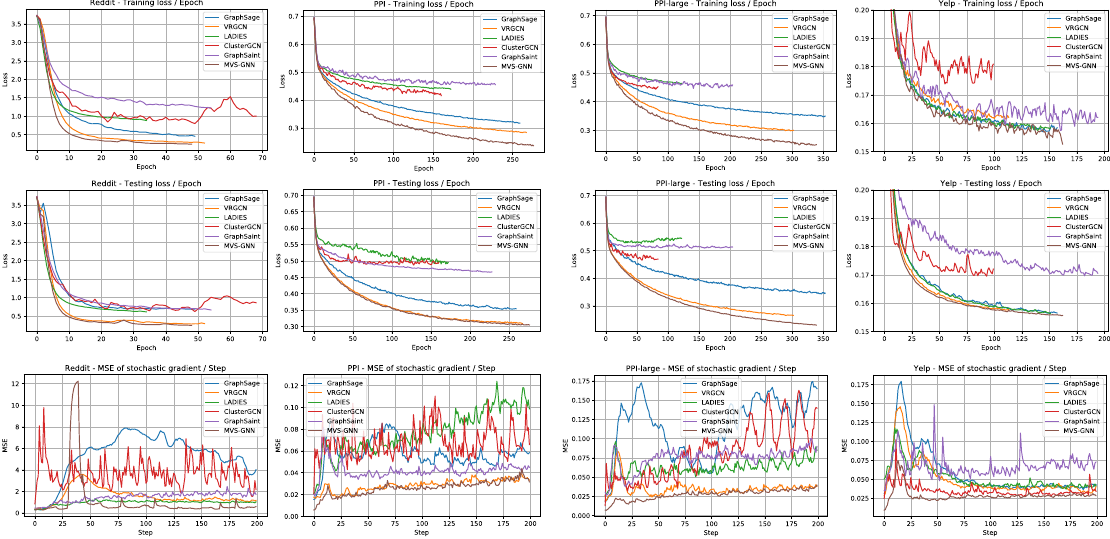}
    \caption{Convergence curves and gradient variance of 2-layer \texttt{MVS-GNN} and baseline models on \texttt{Reddit}, \texttt{PPI}, \texttt{PPI-large}, and \texttt{Yelp} dataset with batch size 512.} 
  \label{fig:loss_grad_norm_flickr}
\end{figure*}

\begin{table*}[t]
\caption{Dataset statistics. \textbf{s} and  \textbf{m} stand for single and \textbf{m}ulti-class classification problems, respectively.}
\label{table:datasets}
\centering
\begin{tabular}{|r|c|c|c|c|c|c|}
\hline
\texttt{Dataset}   & \textbf{Nodes} & \textbf{Edges} & \textbf{Degree} & \textbf{Feature} & \textbf{Classes} & \textbf{Train/Val/Test} \\ \hline\hline
\texttt{Reddit}    & 232,965        & 11,606,919     & 50              & 602              & 41(\textbf{s})   & 66\%/10\%/24\%          \\ \hline
\texttt{PPI}       & 14,755         & 225,270        & 15              & 50               & 121(\textbf{m})  & 66\%/12\%/22\%          \\ \hline
\texttt{PPI-large} & 56,944         & 2,818,716      & 14              & 50               & 121(\textbf{m})  & 79\%/11\%/10\%          \\ \hline
\texttt{Yelp}      & 716,847        & 6,977,410      & 10              & 300              & 100(\textbf{m})  & 75\%/10\%/15\%          \\ \hline
\end{tabular}
\end{table*}

\noindent\textbf{Experiment setup.~} Experiments are under semi-supervised learning setting. We evaluate on the following real-world datasets: (1) \texttt{Reddit}: classifying communities of online posts based on user comments; (2) \texttt{PPI} and \texttt{PPI-large} : classifying protein functions based on the interactions of human tissue proteins; (3) \texttt{Yelp}: classifying product categories based on customer reviewers and friendship.
Detailed information are summarised in Table \ref{table:datasets}.

We compare with five baselines: node-wise sampling methods \texttt{GraphSage} and \texttt{VRGCN}, a layer-wise sampling method \texttt{LADIES}, and subgraph sampling methods \texttt{ClusterGCN} and \texttt{GraphSaint}. 
For a given dataset, we keep the GNN structure the same across all methods. 
We train GNN with the default Laplacian multiplication aggregation defined in \cite{kipf2016semi} for \texttt{Reddit} dataset 
\begin{equation*}
    \mathbf{H}_i^{(\ell)} = \sigma\left(\sum_{j\in\mathcal{N}(i)} L_{i,j} \mathbf{H}_j^{(\ell-1)} \mathbf{W}^{(\ell)}\right),
\end{equation*}
and add an extra concatenate operation defined in \cite{hamilton2017inductive} for \texttt{PPI}, \texttt{PPI-large}, and \texttt{Yelp} datasets. We train GNN with the default Laplacian multiplication aggregation defined in \cite{kipf2016semi} for \texttt{Reddit} dataset
\begin{equation*}
    \mathbf{H}_i^{(\ell)} = \sigma\left(\text{concat}\left(\mathbf{H}_i^{(\ell-1)}, \sum_{j\in\mathcal{N}(i)} L_{i,j} \mathbf{H}_j^{(\ell-1)}\right) \mathbf{W}^{(\ell)}\right).
\end{equation*}
We make this decision because the default Laplacian multiplication aggregation is prone to diverge on multi-class classification dataset. 


By default, we train 2-layer GNNs with hidden state dimension as $F = 256$.
For node-wise sampling methods, we chose 
$5$ neighbors to be sampled for \texttt{GraphSage} and $2$ neighbors to be sampled for \texttt{VRGCN}.
For the layer-wise sampling method, we choose the layer node sample size the same as the current batch size for \texttt{LADIES} (e.g., if the mini-batch size is $512$, then the layer node sample size also equals to $512$ nodes).
For the subgraph sampling method, we partition a graph into clusters of size $128$ and construct the mini-batch by choosing the desired number of clusters for \texttt{ClusterGCN}, and choose node sampling method for \texttt{GraphSaint}.
We chose the checkpoint sampling ratio ($\gamma$) $10\%$ for \texttt{Reddit}, $100\%$ for \texttt{PPI}, $20\%$ for \texttt{PPI-large}, and $2\%$ for \texttt{Yelp} dataset.
All methods terminate when the validation accuracy does not increase a threshold $0.01$ for $400$ mini-batches on \texttt{Reddit}, \texttt{Yelp} datasets and $1000$ mini-batches on \texttt{PPI} and \texttt{PPI-large} datasets. We conduct training for $3$ times and take the mean of the evaluation results.
We choose inner-loop size $K=20$ as default and update the model with Adam optimizer with a learning rate of $0.01$.

\begin{table}[t]
\caption{Comparison of test set F1-micro for  various batch sizes. $\P$ stands for out of memory error. }
\label{table:compare_f1_with_bsize}
\centering
\begin{tabular}{|l|l|c|c|c|c|}
\hline
\textbf{\begin{tabular}[c]{@{}l@{}}Batch \\ Size \end{tabular}}            & \textbf{Method}     & \textbf{Reddit} & \textbf{PPI} & \textbf{PPI-large} & \textbf{Yelp}\\ \hline\hline
\multirow{6}{*}{\textbf{256}}  & \texttt{MVS-GNN}    & \bf{0.938}                & \bf{0.836}            & \bf{0.841}                  & \bf{0.613} \\ \cline{2-6} 
                               & \texttt{GraphSage}  & 0.920                     & 0.739                 & 0.660                       & 0.589 \\ \cline{2-6} 
                               & \texttt{VRGCN}      & 0.917                     & 0.812                 & 0.821                       & 0.555 \\ \cline{2-6} 
                               & \texttt{LADIES}     & 0.932                     & 0.583                 & 0.603                       & 0.596 \\ \cline{2-6} 
                               & \texttt{ClusterGCN} & 0.739                     & 0.586                 & 0.608                       & 0.538 \\ \cline{2-6} 
                               & \texttt{GraphSaint} & 0.907                     & 0.506                 & 0.427                       & 0.514 \\ \hline\hline
\multirow{6}{*}{\textbf{512}}  & \texttt{MVS-GNN}    & \bf{0.942}                & \bf{0.859}            & \bf{0.864}                  & \bf{0.617}\\ \cline{2-6} 
                               & \texttt{GraphSage}  & 0.932                     & 0.781                 & 0.766                       & 0.606\\ \cline{2-6} 
                               & \texttt{VRGCN}      & 0.929                     & 0.831                 & 0.829                       & 0.607 \\ \cline{2-6} 
                               & \texttt{LADIES}     & 0.938                     & 0.607                 & 0.600                       & 0.596\\ \cline{2-6} 
                               & \texttt{ClusterGCN} & 0.897                     & 0.590                 & 0.605                       & 0.577\\ \cline{2-6} 
                               & \texttt{GraphSaint} & 0.921                     & 0.577                 & 0.531                       & 0.540\\ \hline\hline
\multirow{6}{*}{\textbf{1024}} & \texttt{MVS-GNN}    & \bf{0.946}                & \bf{0.864}            & \bf{0.875}                  & \bf{0.619} \\ \cline{2-6} 
                               & \texttt{GraphSage}  & 0.939                     & 0.809                 & 0.789                       & 0.611 \\ \cline{2-6} 
                               & \texttt{VRGCN}      & 0.934                     & 0.848                 & 0.849                       & 0.615 \\ \cline{2-6} 
                               & \texttt{LADIES}     & 0.937                     & 0.659                 & 0.599                       & 0.599 \\ \cline{2-6} 
                               & \texttt{ClusterGCN} & 0.923                     & 0.587                 & 0.639                       & 0.595 \\ \cline{2-6} 
                               & \texttt{GraphSaint} & 0.929                     & 0.611                 & 0.558                       & 0.550 \\ \hline\hline
\multirow{6}{*}{\textbf{2048}} & \texttt{MVS-GNN}    & \bf{0.949}                & \bf{0.880}            & \bf{0.892}                  & \bf{0.620} \\ \cline{2-6}  
                               & \texttt{GraphSage}  & 0.944                     & 0.839                 & 0.833                       & 0.617 \\ \cline{2-6} 
                               & \texttt{VRGCN}      & 0.945                     & 0.844                 & 0.856                       & $\P$ \\ \cline{2-6} 
                               & \texttt{LADIES}     & 0.943                     & 0.722                 & 0.623                       & 0.602 \\ \cline{2-6} 
                               & \texttt{ClusterGCN} & 0.939                     & 0.592                 & 0.647                       & 0.616 \\ \cline{2-6} 
                               & \texttt{GraphSaint} & 0.931                     & 0.633                 & 0.593                       & 0.559 \\ \hline
\end{tabular}
\end{table}

\noindent\textbf{The effect of mini-batch size.~}
Table \ref{table:compare_f1_with_bsize} shows the accuracy comparison of various methods using different batch sizes. 
Clearly, with decoupled variance reduction, \texttt{MVS-GNN} achieves significantly higher accuracy, even when the batch size is small. 
Compared with \texttt{VRGCN}, since \texttt{MVS-GNN} has ``free'' and ``up-to-date'' full-batch history activations every $K$ iterations, this guarantees the effectiveness of function value variance reduction of our model during training. 
Compared with \texttt{GraphSaint} and \texttt{ClusterGCN}, \texttt{GraphSaint} performs node-wise graph sampling, which leads to a sparse small graph with high variance when batch size is small, while \texttt{ClusterGCN} first partition graph into several clusters and construct a dense small graph which is highly biased when the batch size is small.


\noindent\textbf{Effectiveness of variance reduction.~}
Figure \ref{fig:loss_grad_norm_flickr} shows the mean-square error of stochastic gradient and convergence of various methods.
Clearly, minimal variance sampling can lead to a variance reduction of mini-batch estimated gradient and has a positive effect on model performance.


\begin{table}[t]
\caption{Comparison of average time (seconds) on \texttt{PPI} dataset for 5-layer GNN with batch size $512$.}
\label{table:time}
\centering
\begin{tabular}{|l|l|l|l|l|}
\hline
\textbf{Method}     & $T_\text{Sample}$ & $T_\text{Train}$  & $T_\text{Dists}$ & $T_\text{total}$ \\ \hline\hline
\texttt{MVS-GNN}    &  1.057 & 0.646 & 0.088 & 1.791 \\ \hline
\texttt{GraphSage}  &  9.737 & 0.688 & 0     & 10.425\\ \hline
\texttt{VRGCN}      &  10.095& 1.038 & 0     & 11.133\\ \hline
\texttt{LADIES}     &  1.031 & 0.295 & 0     & 1.326 \\ \hline
\texttt{ClusterGCN} &  1.140 & 0.672 & 0     & 1.812 \\ \hline
\texttt{GraphSaint} &  0.793 & 0.214 & 0     & 1.007\\ \hline
\end{tabular}
\end{table}

\noindent\textbf{Evaluation of total time.~} Table \ref{table:time} shows the comparison of time $T_\text{Sample}$, $T_\text{Train}$, $T_\text{Dists}$ on \texttt{PPI} dataset. $T_\text{Sample}$ is defined as the time of constructing $20$ mini-batches for training (in \texttt{MVS-GNN} is the time of $1$ large-batch and $19$ mini-batches). $T_\text{Train}$ is defined as the time to run $20$ mini-batches for training (in \texttt{MVS-GNN} is the time of $1$ large-batch and $19$ mini-batches). $T_\text{Dists}$ is defined as the time to calculate the importance sampling distribution of each node for minimal variance sampling. Therefore, the total time for $20$ iterations is $T_\text{total}=T_\text{Sample}+T_\text{Train}+T_\text{Dists}$.
To achieve fair comparison in terms of sampling complexity, we implement all sampling methods using Python \texttt{scipy.sparse} and \texttt{numpy.random} package, and construct $20$ mini-batches in parallel by Python \texttt{multiprocessing} package with $10$ threads. We choose the default setup and calculate the sample distribution every $20$ iterations for \texttt{MVS-GNN} with importance sampling ratio $100\%$.
Because our method does not need to recursively sample neighbors for each layer and each node in the mini-batch, less time is required. Besides, since a constant number of nodes are calculated in each layer, our method is exponentially faster than node-wise sampling algorithms with respect to the number of layers. 

\begin{figure}
    \centering
    \includegraphics[width=.49\textwidth]{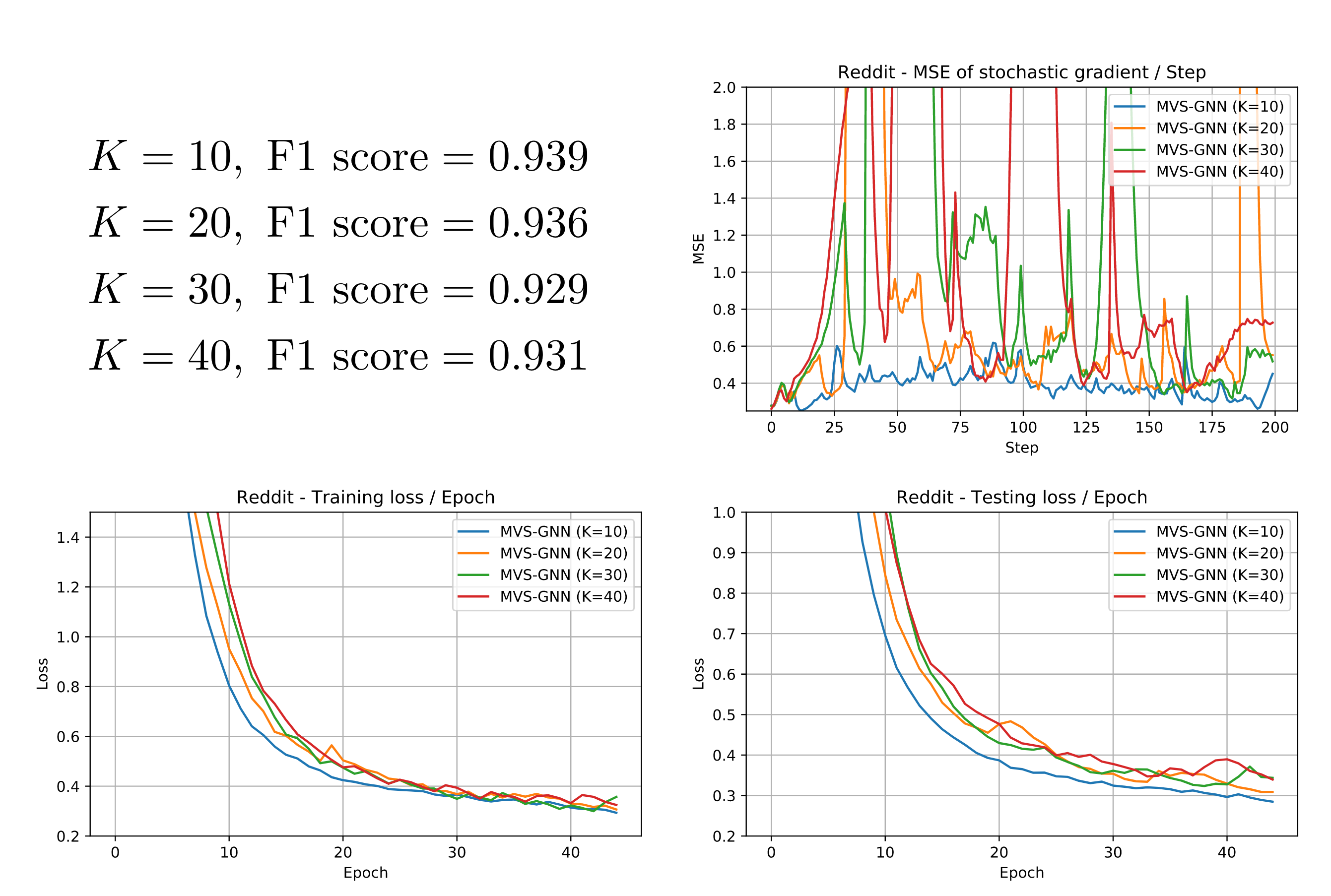}
    \caption{Comparison of gradient variance, training loss, and testing loss on \texttt{Reddit} dataset with different number of inner-loop iterations ($K=10,20,30,40$).}
    \label{fig:grad_freq}
\end{figure}

\noindent\textbf{Evaluation on inner-loop interval.~}
\texttt{MVS-GNN} requires performing large-batch training periodically to calculate the importance sampling distribution. A larger number of inner-loop interval ($K$) can make training speed faster, but also might make the importance sample distribution too stale to represent the true distribution. 
In Figure \ref{fig:grad_freq} , we show the comparison of gradient variance, training loss, and testing loss with different number of inner-loop intervals on \texttt{Reddit} dataset. We choose mini-batch size $512$, dropout rate $0.1$, importance sampling ratio $10\%$, and change the inner-loop intervals from $10$ mini-batches to $30$ mini-batches. 

\begin{figure*}[t]
    \centering
    \includegraphics[width=0.7\textwidth]{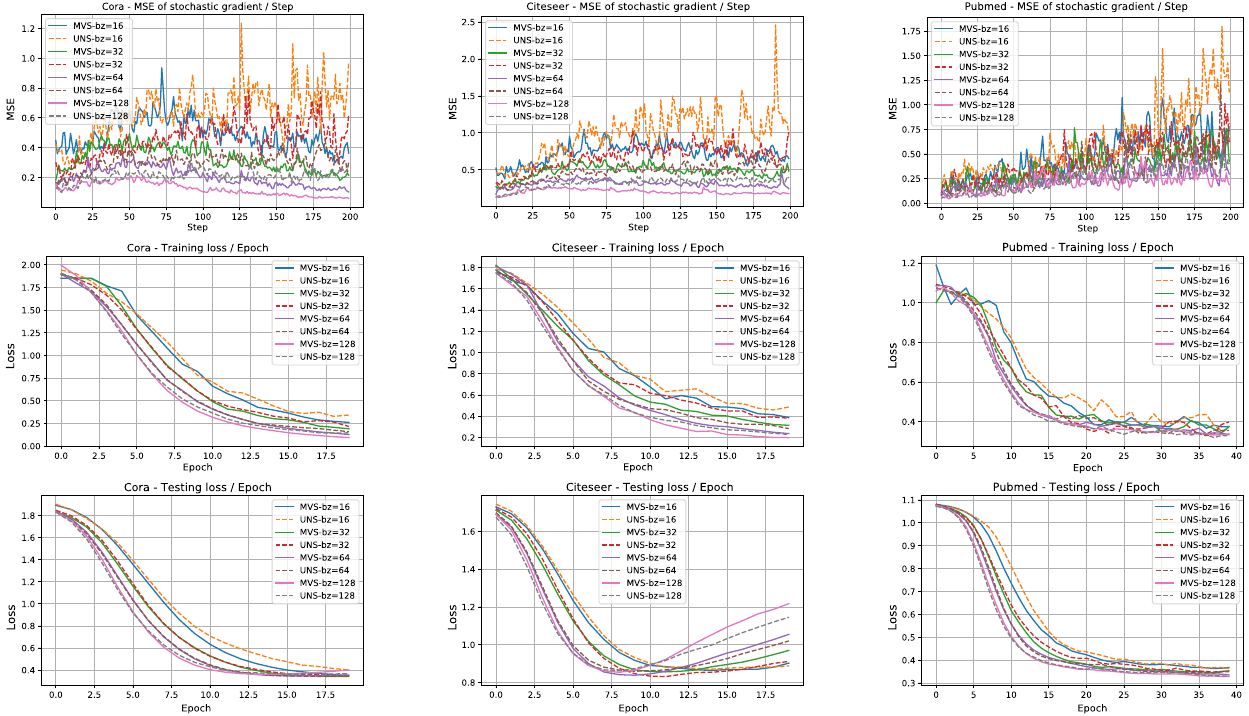}
    \caption{Comparison of gradient variance, training loss, and testing loss with small mini-batch size on \texttt{Cora}, \texttt{Citeseer}, and \texttt{Pubmed} datasets.}
    \label{fig:extreme_small_bsize}
\end{figure*}

\noindent\textbf{Evaluation on small mini-batch size.~}
In Figure \ref{fig:extreme_small_bsize}, we show the effectiveness of minimal variance sampling using small mini-batch size on \texttt{Cora}, \texttt{Citeseer}, and \texttt{Pubmed} dataset introduce in \cite{kipf2016semi}. To eliminate the embedding approximation variance, we use all neighbors to inference the embedding matrix, such that the only randomness happens at choosing nodes in mini-batch, which is the original intention minimal variance sampling designed for. We choose importance sampling ratio as $50\%$ for Pubmed, $100\%$ for Cora and Citeseer, and update the importance sampling distribution every $10$ iterations (shown as $1$ epoch in Figure~\ref{fig:extreme_small_bsize}). We choose hidden state as $64$, dropout ratio as $0.1$, change the mini-batch size (\emph{bz}), and monitor the difference of gradient variance, training loss, and testing loss between minimal variance sampling (MVS) and uniform sampling (UNS). Our result shows that minimal variance sampling can significantly reduce the gradient variance and accelerate the convergence speed during training.

\section{Conclusion}\label{section:conclusion}
In this work, we theoretically analyzed the variance of sampling based methods for training GCNs and demonstrated  that, due to  composite structure of empirical risk, the variance of any sampling method can be decomposed as embedding approximation variance and stochastic gradient variance. To mitigate these two types of variance and obtain faster convergence, a decoupled variance reduction strategy is proposed that employs gradient information to sample nodes with minimal variance and explicitly reduce the variance introduced by embedding approximation. We empirically demonstrate  the superior performance of the proposed decoupled variance reduction method in comparison  with the exiting sampling methods, where it enjoys a faster convergence rate and a better  generalization performance   even with smaller mini-batch sizes. We leave exploring the empirical efficiency of proposed methods to other variants of GNNs such as graph classification and attention based GNNs as a future study.

\bibliographystyle{ACM-Reference-Format}
\bibliography{sample-base}


\begin{thebibliography}{33}


\ifx \showCODEN    \undefined \def \showCODEN     #1{\unskip}     \fi
\ifx \showDOI      \undefined \def \showDOI       #1{#1}\fi
\ifx \showISBNx    \undefined \def \showISBNx     #1{\unskip}     \fi
\ifx \showISBNxiii \undefined \def \showISBNxiii  #1{\unskip}     \fi
\ifx \showISSN     \undefined \def \showISSN      #1{\unskip}     \fi
\ifx \showLCCN     \undefined \def \showLCCN      #1{\unskip}     \fi
\ifx \shownote     \undefined \def \shownote      #1{#1}          \fi
\ifx \showarticletitle \undefined \def \showarticletitle #1{#1}   \fi
\ifx \showURL      \undefined \def \showURL       {\relax}        \fi
\providecommand\bibfield[2]{#2}
\providecommand\bibinfo[2]{#2}
\providecommand\natexlab[1]{#1}
\providecommand\showeprint[2][]{arXiv:#2}

\bibitem[\protect\citeauthoryear{Abadi, Barham, Chen, Chen, Davis, Dean, Devin,
  Ghemawat, Irving, Isard, et~al\mbox{.}}{Abadi et~al\mbox{.}}{2016}]%
        {abadi2016tensorflow}
\bibfield{author}{\bibinfo{person}{Mart{\'\i}n Abadi}, \bibinfo{person}{Paul
  Barham}, \bibinfo{person}{Jianmin Chen}, \bibinfo{person}{Zhifeng Chen},
  \bibinfo{person}{Andy Davis}, \bibinfo{person}{Jeffrey Dean},
  \bibinfo{person}{Matthieu Devin}, \bibinfo{person}{Sanjay Ghemawat},
  \bibinfo{person}{Geoffrey Irving}, \bibinfo{person}{Michael Isard},
  {et~al\mbox{.}}} \bibinfo{year}{2016}\natexlab{}.
\newblock \showarticletitle{Tensorflow: A system for large-scale machine
  learning}. In \bibinfo{booktitle}{\emph{12th $\{$USENIX$\}$ Symposium on
  Operating Systems Design and Implementation ($\{$OSDI$\}$ 16)}}.
  \bibinfo{pages}{265--283}.
\newblock


\bibitem[\protect\citeauthoryear{Berg, Kipf, and Welling}{Berg
  et~al\mbox{.}}{2017}]%
        {berg2017graph}
\bibfield{author}{\bibinfo{person}{Rianne van~den Berg},
  \bibinfo{person}{Thomas~N Kipf}, {and} \bibinfo{person}{Max Welling}.}
  \bibinfo{year}{2017}\natexlab{}.
\newblock \showarticletitle{Graph convolutional matrix completion}.
\newblock \bibinfo{journal}{\emph{arXiv preprint arXiv:1706.02263}}
  (\bibinfo{year}{2017}).
\newblock


\bibitem[\protect\citeauthoryear{Chen, Ma, and Xiao}{Chen
  et~al\mbox{.}}{2018}]%
        {chen2018fastgcn}
\bibfield{author}{\bibinfo{person}{Jie Chen}, \bibinfo{person}{Tengfei Ma},
  {and} \bibinfo{person}{Cao Xiao}.} \bibinfo{year}{2018}\natexlab{}.
\newblock \showarticletitle{Fastgcn: fast learning with graph convolutional
  networks via importance sampling}.
\newblock \bibinfo{journal}{\emph{arXiv preprint arXiv:1801.10247}}
  (\bibinfo{year}{2018}).
\newblock


\bibitem[\protect\citeauthoryear{Chen, Zhu, and Song}{Chen
  et~al\mbox{.}}{2017}]%
        {chen2017stochastic}
\bibfield{author}{\bibinfo{person}{Jianfei Chen}, \bibinfo{person}{Jun Zhu},
  {and} \bibinfo{person}{Le Song}.} \bibinfo{year}{2017}\natexlab{}.
\newblock \showarticletitle{Stochastic training of graph convolutional networks
  with variance reduction}.
\newblock \bibinfo{journal}{\emph{arXiv preprint arXiv:1710.10568}}
  (\bibinfo{year}{2017}).
\newblock


\bibitem[\protect\citeauthoryear{Chiang, Liu, Si, Li, Bengio, and Hsieh}{Chiang
  et~al\mbox{.}}{2019}]%
        {chiang2019cluster}
\bibfield{author}{\bibinfo{person}{Wei-Lin Chiang}, \bibinfo{person}{Xuanqing
  Liu}, \bibinfo{person}{Si Si}, \bibinfo{person}{Yang Li},
  \bibinfo{person}{Samy Bengio}, {and} \bibinfo{person}{Cho-Jui Hsieh}.}
  \bibinfo{year}{2019}\natexlab{}.
\newblock \showarticletitle{Cluster-gcn: An efficient algorithm for training
  deep and large graph convolutional networks}. In
  \bibinfo{booktitle}{\emph{Proceedings of the 25th ACM SIGKDD International
  Conference on Knowledge Discovery \& Data Mining}}.
  \bibinfo{pages}{257--266}.
\newblock


\bibitem[\protect\citeauthoryear{Csiba, Qu, and Richt{\'a}rik}{Csiba
  et~al\mbox{.}}{2015}]%
        {csiba2015stochastic}
\bibfield{author}{\bibinfo{person}{Dominik Csiba}, \bibinfo{person}{Zheng Qu},
  {and} \bibinfo{person}{Peter Richt{\'a}rik}.}
  \bibinfo{year}{2015}\natexlab{}.
\newblock \showarticletitle{Stochastic dual coordinate ascent with adaptive
  probabilities}. In \bibinfo{booktitle}{\emph{ICML}}.
  \bibinfo{pages}{674--683}.
\newblock


\bibitem[\protect\citeauthoryear{Cui, Henrickson, Ke, and Wang}{Cui
  et~al\mbox{.}}{2019}]%
        {cui2019traffic}
\bibfield{author}{\bibinfo{person}{Zhiyong Cui}, \bibinfo{person}{Kristian
  Henrickson}, \bibinfo{person}{Ruimin Ke}, {and} \bibinfo{person}{Yinhai
  Wang}.} \bibinfo{year}{2019}\natexlab{}.
\newblock \showarticletitle{Traffic graph convolutional recurrent neural
  network: A deep learning framework for network-scale traffic learning and
  forecasting}.
\newblock \bibinfo{journal}{\emph{IEEE Transactions on Intelligent
  Transportation Systems}} (\bibinfo{year}{2019}).
\newblock


\bibitem[\protect\citeauthoryear{Deng, Rangwala, and Ning}{Deng
  et~al\mbox{.}}{2019}]%
        {deng2019learning}
\bibfield{author}{\bibinfo{person}{Songgaojun Deng}, \bibinfo{person}{Huzefa
  Rangwala}, {and} \bibinfo{person}{Yue Ning}.}
  \bibinfo{year}{2019}\natexlab{}.
\newblock \showarticletitle{Learning Dynamic Context Graphs for Predicting
  Social Events}. In \bibinfo{booktitle}{\emph{KDD}}.
  \bibinfo{pages}{1007--1016}.
\newblock


\bibitem[\protect\citeauthoryear{Do, Tran, and Venkatesh}{Do
  et~al\mbox{.}}{2019}]%
        {do2019graph}
\bibfield{author}{\bibinfo{person}{Kien Do}, \bibinfo{person}{Truyen Tran},
  {and} \bibinfo{person}{Svetha Venkatesh}.} \bibinfo{year}{2019}\natexlab{}.
\newblock \showarticletitle{Graph transformation policy network for chemical
  reaction prediction}. In \bibinfo{booktitle}{\emph{KDD}}.
  \bibinfo{pages}{750--760}.
\newblock


\bibitem[\protect\citeauthoryear{Duvenaud, Maclaurin, Iparraguirre, Bombarell,
  Hirzel, Aspuru-Guzik, and Adams}{Duvenaud et~al\mbox{.}}{2015}]%
        {duvenaud2015convolutional}
\bibfield{author}{\bibinfo{person}{David~K Duvenaud}, \bibinfo{person}{Dougal
  Maclaurin}, \bibinfo{person}{Jorge Iparraguirre}, \bibinfo{person}{Rafael
  Bombarell}, \bibinfo{person}{Timothy Hirzel}, \bibinfo{person}{Al{\'a}n
  Aspuru-Guzik}, {and} \bibinfo{person}{Ryan~P Adams}.}
  \bibinfo{year}{2015}\natexlab{}.
\newblock \showarticletitle{Convolutional networks on graphs for learning
  molecular fingerprints}. In \bibinfo{booktitle}{\emph{NeurIPS}}.
  \bibinfo{pages}{2224--2232}.
\newblock


\bibitem[\protect\citeauthoryear{Hamilton, Ying, and Leskovec}{Hamilton
  et~al\mbox{.}}{2017}]%
        {hamilton2017inductive}
\bibfield{author}{\bibinfo{person}{Will Hamilton}, \bibinfo{person}{Zhitao
  Ying}, {and} \bibinfo{person}{Jure Leskovec}.}
  \bibinfo{year}{2017}\natexlab{}.
\newblock \showarticletitle{Inductive representation learning on large graphs}.
  In \bibinfo{booktitle}{\emph{NeurIPS}}. \bibinfo{pages}{1024--1034}.
\newblock


\bibitem[\protect\citeauthoryear{Katharopoulos and Fleuret}{Katharopoulos and
  Fleuret}{2018}]%
        {katharopoulos2018not}
\bibfield{author}{\bibinfo{person}{Angelos Katharopoulos} {and}
  \bibinfo{person}{Fran{\c{c}}ois Fleuret}.} \bibinfo{year}{2018}\natexlab{}.
\newblock \showarticletitle{Not all samples are created equal: Deep learning
  with importance sampling}.
\newblock \bibinfo{journal}{\emph{arXiv preprint arXiv:1803.00942}}
  (\bibinfo{year}{2018}).
\newblock


\bibitem[\protect\citeauthoryear{Kipf and Welling}{Kipf and Welling}{2016}]%
        {kipf2016semi}
\bibfield{author}{\bibinfo{person}{Thomas~N Kipf} {and} \bibinfo{person}{Max
  Welling}.} \bibinfo{year}{2016}\natexlab{}.
\newblock \showarticletitle{Semi-supervised classification with graph
  convolutional networks}.
\newblock \bibinfo{journal}{\emph{arXiv preprint arXiv:1609.02907}}
  (\bibinfo{year}{2016}).
\newblock


\bibitem[\protect\citeauthoryear{Kumar, Zhang, and Leskovec}{Kumar
  et~al\mbox{.}}{2019}]%
        {kumar2019predicting}
\bibfield{author}{\bibinfo{person}{Srijan Kumar}, \bibinfo{person}{Xikun
  Zhang}, {and} \bibinfo{person}{Jure Leskovec}.}
  \bibinfo{year}{2019}\natexlab{}.
\newblock \showarticletitle{Predicting dynamic embedding trajectory in temporal
  interaction networks}. In \bibinfo{booktitle}{\emph{KDD}}.
  \bibinfo{pages}{1269--1278}.
\newblock


\bibitem[\protect\citeauthoryear{Li, Han, Cheng, Su, Wang, Zhang, and Pan}{Li
  et~al\mbox{.}}{2019}]%
        {li2019predicting}
\bibfield{author}{\bibinfo{person}{Jia Li}, \bibinfo{person}{Zhichao Han},
  \bibinfo{person}{Hong Cheng}, \bibinfo{person}{Jiao Su},
  \bibinfo{person}{Pengyun Wang}, \bibinfo{person}{Jianfeng Zhang}, {and}
  \bibinfo{person}{Lujia Pan}.} \bibinfo{year}{2019}\natexlab{}.
\newblock \showarticletitle{Predicting Path Failure In Time-Evolving Graphs}.
  In \bibinfo{booktitle}{\emph{KDD}}. \bibinfo{pages}{1279--1289}.
\newblock


\bibitem[\protect\citeauthoryear{Li, Wang, Zhu, and Huang}{Li
  et~al\mbox{.}}{2018}]%
        {li2018adaptive}
\bibfield{author}{\bibinfo{person}{Ruoyu Li}, \bibinfo{person}{Sheng Wang},
  \bibinfo{person}{Feiyun Zhu}, {and} \bibinfo{person}{Junzhou Huang}.}
  \bibinfo{year}{2018}\natexlab{}.
\newblock \showarticletitle{Adaptive graph convolutional neural networks}. In
  \bibinfo{booktitle}{\emph{AAAI}}.
\newblock


\bibitem[\protect\citeauthoryear{Papa, Bianchi, and Cl{\'e}men{\c{c}}on}{Papa
  et~al\mbox{.}}{2015}]%
        {papa2015adaptive}
\bibfield{author}{\bibinfo{person}{Guillaume Papa}, \bibinfo{person}{Pascal
  Bianchi}, {and} \bibinfo{person}{St{\'e}phan Cl{\'e}men{\c{c}}on}.}
  \bibinfo{year}{2015}\natexlab{}.
\newblock \showarticletitle{Adaptive sampling for incremental optimization
  using stochastic gradient descent}. In \bibinfo{booktitle}{\emph{ALT}}.
  Springer, \bibinfo{pages}{317--331}.
\newblock


\bibitem[\protect\citeauthoryear{Park, Kan, Dong, Zhao, and Faloutsos}{Park
  et~al\mbox{.}}{2019}]%
        {park2019estimating}
\bibfield{author}{\bibinfo{person}{Namyong Park}, \bibinfo{person}{Andrey Kan},
  \bibinfo{person}{Xin~Luna Dong}, \bibinfo{person}{Tong Zhao}, {and}
  \bibinfo{person}{Christos Faloutsos}.} \bibinfo{year}{2019}\natexlab{}.
\newblock \showarticletitle{Estimating node importance in knowledge graphs
  using graph neural networks}. In \bibinfo{booktitle}{\emph{KDD}}.
  \bibinfo{pages}{596--606}.
\newblock


\bibitem[\protect\citeauthoryear{Paszke, Gross, Massa, Lerer, Bradbury, Chanan,
  Killeen, Lin, Gimelshein, Antiga, et~al\mbox{.}}{Paszke
  et~al\mbox{.}}{2019}]%
        {paszke2019pytorch}
\bibfield{author}{\bibinfo{person}{Adam Paszke}, \bibinfo{person}{Sam Gross},
  \bibinfo{person}{Francisco Massa}, \bibinfo{person}{Adam Lerer},
  \bibinfo{person}{James Bradbury}, \bibinfo{person}{Gregory Chanan},
  \bibinfo{person}{Trevor Killeen}, \bibinfo{person}{Zeming Lin},
  \bibinfo{person}{Natalia Gimelshein}, \bibinfo{person}{Luca Antiga},
  {et~al\mbox{.}}} \bibinfo{year}{2019}\natexlab{}.
\newblock \showarticletitle{PyTorch: An imperative style, high-performance deep
  learning library}. In \bibinfo{booktitle}{\emph{NeurIPS}}.
  \bibinfo{pages}{8024--8035}.
\newblock


\bibitem[\protect\citeauthoryear{Qiu, Tang, Ma, Dong, Wang, and Tang}{Qiu
  et~al\mbox{.}}{2018}]%
        {qiu2018deepinf}
\bibfield{author}{\bibinfo{person}{Jiezhong Qiu}, \bibinfo{person}{Jian Tang},
  \bibinfo{person}{Hao Ma}, \bibinfo{person}{Yuxiao Dong},
  \bibinfo{person}{Kuansan Wang}, {and} \bibinfo{person}{Jie Tang}.}
  \bibinfo{year}{2018}\natexlab{}.
\newblock \showarticletitle{DeepInf: Modeling influence locality in large
  social networks}. In \bibinfo{booktitle}{\emph{KDD}}.
\newblock


\bibitem[\protect\citeauthoryear{Rahimi, Cohn, and Baldwin}{Rahimi
  et~al\mbox{.}}{2018}]%
        {rahimi2018semi}
\bibfield{author}{\bibinfo{person}{Afshin Rahimi}, \bibinfo{person}{Trevor
  Cohn}, {and} \bibinfo{person}{Timothy Baldwin}.}
  \bibinfo{year}{2018}\natexlab{}.
\newblock \showarticletitle{Semi-supervised user geolocation via graph
  convolutional networks}.
\newblock \bibinfo{journal}{\emph{arXiv preprint arXiv:1804.08049}}
  (\bibinfo{year}{2018}).
\newblock


\bibitem[\protect\citeauthoryear{Salehi, Thiran, and Celis}{Salehi
  et~al\mbox{.}}{2018}]%
        {salehi2018coordinate}
\bibfield{author}{\bibinfo{person}{Farnood Salehi}, \bibinfo{person}{Patrick
  Thiran}, {and} \bibinfo{person}{Elisa Celis}.}
  \bibinfo{year}{2018}\natexlab{}.
\newblock \showarticletitle{Coordinate descent with bandit sampling}. In
  \bibinfo{booktitle}{\emph{NeurIPS}}. \bibinfo{pages}{9247--9257}.
\newblock


\bibitem[\protect\citeauthoryear{Stich, Raj, and Jaggi}{Stich
  et~al\mbox{.}}{2017}]%
        {stich2017safe}
\bibfield{author}{\bibinfo{person}{Sebastian~U Stich}, \bibinfo{person}{Anant
  Raj}, {and} \bibinfo{person}{Martin Jaggi}.} \bibinfo{year}{2017}\natexlab{}.
\newblock \showarticletitle{Safe adaptive importance sampling}. In
  \bibinfo{booktitle}{\emph{NeurIPS}}. \bibinfo{pages}{4381--4391}.
\newblock


\bibitem[\protect\citeauthoryear{Wang, Xu, Liu, Lian, Chen, Du, Wu, and
  Su}{Wang et~al\mbox{.}}{2019b}]%
        {wang2019mcne}
\bibfield{author}{\bibinfo{person}{Hao Wang}, \bibinfo{person}{Tong Xu},
  \bibinfo{person}{Qi Liu}, \bibinfo{person}{Defu Lian},
  \bibinfo{person}{Enhong Chen}, \bibinfo{person}{Dongfang Du},
  \bibinfo{person}{Han Wu}, {and} \bibinfo{person}{Wen Su}.}
  \bibinfo{year}{2019}\natexlab{b}.
\newblock \showarticletitle{MCNE: An End-to-End Framework for Learning Multiple
  Conditional Network Representations of Social Network}. In
  \bibinfo{booktitle}{\emph{KDD}}. \bibinfo{pages}{1064--1072}.
\newblock


\bibitem[\protect\citeauthoryear{Wang, Zhang, Zhang, Leskovec, Zhao, Li, and
  Wang}{Wang et~al\mbox{.}}{2019c}]%
        {wang2019knowledge}
\bibfield{author}{\bibinfo{person}{Hongwei Wang}, \bibinfo{person}{Fuzheng
  Zhang}, \bibinfo{person}{Mengdi Zhang}, \bibinfo{person}{Jure Leskovec},
  \bibinfo{person}{Miao Zhao}, \bibinfo{person}{Wenjie Li}, {and}
  \bibinfo{person}{Zhongyuan Wang}.} \bibinfo{year}{2019}\natexlab{c}.
\newblock \showarticletitle{Knowledge-aware graph neural networks with label
  smoothness regularization for recommender systems}. In
  \bibinfo{booktitle}{\emph{KDD}}. \bibinfo{pages}{968--977}.
\newblock


\bibitem[\protect\citeauthoryear{Wang, He, Cao, Liu, and Chua}{Wang
  et~al\mbox{.}}{2019a}]%
        {wang2019kgat}
\bibfield{author}{\bibinfo{person}{Xiang Wang}, \bibinfo{person}{Xiangnan He},
  \bibinfo{person}{Yixin Cao}, \bibinfo{person}{Meng Liu}, {and}
  \bibinfo{person}{Tat-Seng Chua}.} \bibinfo{year}{2019}\natexlab{a}.
\newblock \showarticletitle{Kgat: Knowledge graph attention network for
  recommendation}. In \bibinfo{booktitle}{\emph{KDD}}.
  \bibinfo{pages}{950--958}.
\newblock


\bibitem[\protect\citeauthoryear{Ying, He, Chen, Eksombatchai, Hamilton, and
  Leskovec}{Ying et~al\mbox{.}}{2018}]%
        {ying2018graph}
\bibfield{author}{\bibinfo{person}{Rex Ying}, \bibinfo{person}{Ruining He},
  \bibinfo{person}{Kaifeng Chen}, \bibinfo{person}{Pong Eksombatchai},
  \bibinfo{person}{William~L Hamilton}, {and} \bibinfo{person}{Jure Leskovec}.}
  \bibinfo{year}{2018}\natexlab{}.
\newblock \showarticletitle{Graph convolutional neural networks for web-scale
  recommender systems}. In \bibinfo{booktitle}{\emph{KDD}}.
  \bibinfo{pages}{974--983}.
\newblock


\bibitem[\protect\citeauthoryear{Zeng, Zhou, Srivastava, Kannan, and
  Prasanna}{Zeng et~al\mbox{.}}{2019}]%
        {zeng2019graphsaint}
\bibfield{author}{\bibinfo{person}{Hanqing Zeng}, \bibinfo{person}{Hongkuan
  Zhou}, \bibinfo{person}{Ajitesh Srivastava}, \bibinfo{person}{Rajgopal
  Kannan}, {and} \bibinfo{person}{Viktor Prasanna}.}
  \bibinfo{year}{2019}\natexlab{}.
\newblock \showarticletitle{Graphsaint: Graph sampling based inductive learning
  method}.
\newblock \bibinfo{journal}{\emph{arXiv preprint arXiv:1907.04931}}
  (\bibinfo{year}{2019}).
\newblock


\bibitem[\protect\citeauthoryear{Zhang, Kjellstrom, and Mandt}{Zhang
  et~al\mbox{.}}{2017}]%
        {zhang2017determinantal}
\bibfield{author}{\bibinfo{person}{Cheng Zhang}, \bibinfo{person}{Hedvig
  Kjellstrom}, {and} \bibinfo{person}{Stephan Mandt}.}
  \bibinfo{year}{2017}\natexlab{}.
\newblock \showarticletitle{Determinantal point processes for mini-batch
  diversification}.
\newblock \bibinfo{journal}{\emph{arXiv preprint arXiv:1705.00607}}
  (\bibinfo{year}{2017}).
\newblock


\bibitem[\protect\citeauthoryear{Zhao and Zhang}{Zhao and Zhang}{2015}]%
        {zhao2015stochastic}
\bibfield{author}{\bibinfo{person}{Peilin Zhao} {and} \bibinfo{person}{Tong
  Zhang}.} \bibinfo{year}{2015}\natexlab{}.
\newblock \showarticletitle{Stochastic optimization with importance sampling
  for regularized loss minimization}. In \bibinfo{booktitle}{\emph{ICML}}.
  \bibinfo{pages}{1--9}.
\newblock


\bibitem[\protect\citeauthoryear{Zheng, Richt{\'a}rik, and Zhang}{Zheng
  et~al\mbox{.}}{2014}]%
        {zheng2014randomized}
\bibfield{author}{\bibinfo{person}{Q Zheng}, \bibinfo{person}{P Richt{\'a}rik},
  {and} \bibinfo{person}{T Zhang}.} \bibinfo{year}{2014}\natexlab{}.
\newblock \bibinfo{title}{Randomized dual coordinate ascent with arbitrary
  sampling}.
\newblock
\newblock


\bibitem[\protect\citeauthoryear{Zhu}{Zhu}{2016}]%
        {zhu2016gradient}
\bibfield{author}{\bibinfo{person}{Rong Zhu}.} \bibinfo{year}{2016}\natexlab{}.
\newblock \showarticletitle{Gradient-based sampling: An adaptive importance
  sampling for least-squares}. In \bibinfo{booktitle}{\emph{NeurIPS}}.
  \bibinfo{pages}{406--414}.
\newblock


\bibitem[\protect\citeauthoryear{Zou, Hu, Wang, Jiang, Sun, and Gu}{Zou
  et~al\mbox{.}}{2019}]%
        {zou2019layer}
\bibfield{author}{\bibinfo{person}{Difan Zou}, \bibinfo{person}{Ziniu Hu},
  \bibinfo{person}{Yewen Wang}, \bibinfo{person}{Song Jiang},
  \bibinfo{person}{Yizhou Sun}, {and} \bibinfo{person}{Quanquan Gu}.}
  \bibinfo{year}{2019}\natexlab{}.
\newblock \showarticletitle{Layer-Dependent Importance Sampling for Training
  Deep and Large Graph Convolutional Networks}. In
  \bibinfo{booktitle}{\emph{NeurIPS}}. \bibinfo{pages}{11247--11256}.
\newblock


\end{thebibliography}

\clearpage
\appendix\onecolumn
\section{Proof of Lemma~\ref{lemma:decouple_bias}}

We can bound $\|\mathbf{g}-\tilde{\mathbf{g}}\|$ by adding and subtracting intermediate terms inside such that each adjacent pair of products differ at most in one factor as follows:
\begin{equation}\label{eq:decouple_bias_1}
    \begin{aligned}
        \cE[\|\mathbf{g}-\tilde{\mathbf{g}}\|^2] &= \cE[\|\nabla f_{\omega_1}^{(1)}(\boldsymbol{\theta}_t) \cdot \nabla f_{\omega_2}^{(2)}(F^{(1)}(\boldsymbol{\theta})) \cdot \nabla f_{\omega_3}^{(2)}(F^{(2)}(\boldsymbol{\theta})) \cdots \nabla f_{\omega_L}^{(L)}(F^{(L-1)}(\boldsymbol{\theta})) \\
        &\qquad - \nabla f^{(1)}_{\omega_1}(\boldsymbol{\theta})\cdot \nabla f^{(2)}_{\omega_2}(f_{\omega_1}^{(1)}(\boldsymbol{\theta})) \cdot \nabla f_{\omega_3}^{(3)}(f_{\omega_2}^{(2)}\circ f_{\omega_1}^{(1)}(\boldsymbol{\theta})) \cdots \nabla f^{(L)}_{\omega_L}(f_{\omega_{L-1}}^{(L-1)}\circ \cdots \circ f_{\omega_1}^{(1)}(\boldsymbol{\theta}))\|^2] \\
        &\leq L\cdot(\cE[\|\nabla f_{\omega_1}^{(1)}(\boldsymbol{\theta}) \cdot \nabla f_{\omega_2}^{(2)}(F^{(1)}(\boldsymbol{\theta})) \cdot \nabla f_{\omega_3}^{(3)}(F^{(2)}(\boldsymbol{\theta})) \cdots \nabla f_{\omega_L}^{(L)}(F^{(L-1)}(\boldsymbol{\theta})) \\
        &\qquad - \nabla f^{(1)}_{\omega_1}(\boldsymbol{\theta}) \cdot \nabla f^{(2)}_{\omega_2}(f_{\omega_1}^{(1)}(\boldsymbol{\theta})) \cdot \nabla f_{\omega_3}^{(3)}(F^{(2)}(\boldsymbol{\theta})) \cdots \nabla f_{\omega_L}^{(L)}(F^{(L-1)}(\boldsymbol{\theta}))\|^2] \\
        &\quad + \cE[\|\nabla f^{(1)}_{\omega_1}(\boldsymbol{\theta}) \cdot \nabla f^{(2)}_{\omega_2}(f_{\omega_1}^{(1)}(\boldsymbol{\theta})) \cdot \nabla f_{\omega_3}^{(3)}(F^{(2)}(\boldsymbol{\theta})) \cdots \nabla f_{\omega_L}^{(L)}(F^{(L-1)}(\boldsymbol{\theta})) \\
        &\qquad - \nabla f^{(1)}_{\omega_1}(\boldsymbol{\theta}) \cdot \nabla f^{(2)}_{\omega_2}(f_{\omega_1}^{(1)}(\boldsymbol{\theta})) \cdot \nabla f_{\omega_3}^{(3)}(f_{\omega_2}^{(2)}\circ f_{\omega_1}^{(1)}(\boldsymbol{\theta})) \cdots \nabla f_{\omega_L}^{(L)}(F^{(L-1)}(\boldsymbol{\theta}))\|^2] +\cdots \\
        &\quad + \cE[\|\nabla f^{(1)}_{\omega_1}(\boldsymbol{\theta})\cdot \nabla f^{(2)}_{\omega_2}(f_{\omega_1}^{(1)}(\boldsymbol{\theta})) \cdot \nabla f_{\omega_3}^{(3)}(f_{\omega_2}^{(2)}\circ f_{\omega_1}^{(1)}(\boldsymbol{\theta})) \cdots \nabla f^{(L)}_{\omega_L}(F^{(L-1)}(\boldsymbol{\theta})) \\
        &\qquad - \nabla f^{(1)}_{\omega_1}(\boldsymbol{\theta})\cdot \nabla f^{(2)}_{\omega_2}(f_{\omega_1}^{(1)}(\boldsymbol{\theta})) \cdot \nabla f_{\omega_3}^{(3)}(f_{\omega_2}^{(2)}\circ f_{\omega_1}^{(1)}(\boldsymbol{\theta})) \cdots \nabla f^{(L)}_{\omega_L}(f_{\omega_{L-1}}^{(L-1)}\circ \cdots \circ f_{\omega_1}^{(1)}(\boldsymbol{\theta}))\|^2] ) \\
        &\leq L\cdot\sum_{\ell=2}^L \left(\sprod_{i=1}^{\ell-1} \rho^2_i \right)\left(\sprod_{i=\ell+1}^L \rho^2_i \right) G^2_\ell \cdot \cE[\|F^{(\ell-1)}(\boldsymbol{\theta}) - f_{\omega_{\ell-1}}^{(\ell-1)}\circ \cdots \circ f_{\omega_1}^{(1)}(\boldsymbol{\theta})\|^2].
    \end{aligned}
\end{equation}

We can bound $\cE[\|F^{(\ell)}(\boldsymbol{\theta}) - f_{\omega_{\ell}}^{(\ell)}\circ \cdots \circ f_{\omega_1}^{(1)}(\boldsymbol{\theta})\|^2]$ by adding and subtracting intermediate terms inside the such that each adjacent pair of products differ at most in one factor.
\begin{equation}\label{eq:decouple_bias_2}
    \begin{aligned}
    \cE[\|F^{(\ell)}(\boldsymbol{\theta}) - f_{\omega_{\ell}}^{(\ell)}\circ \cdots \circ f_{\omega_1}^{(1)}(\boldsymbol{\theta})\|^2] &= \cE[\|f_{\omega_\ell}^{(\ell)} \circ f_{\omega_{\ell-1}}^{(\ell-1)} \circ \cdots \circ f_{\omega_1}^{(1)}(\boldsymbol{\theta}_t) - f^{(\ell)} \circ f^{(\ell-1)}  \circ \cdots\circ f^{(1)}(\boldsymbol{\theta}_t)\|^2] \\
    &\leq \ell \Big(\cE[\|f_{\omega_\ell}^{(\ell)} \circ f_{\omega_{\ell-1}}^{(\ell-1)} \circ \cdots \circ f_{\omega_1}^{(1)}(\boldsymbol{\theta}_t) - f_{\omega_\ell}^{(\ell)} \circ f_{\omega_{\ell-1}}^{(\ell-1)} \circ \cdots \circ f_{\omega_2}^{(2)}(F^{(1)}(\boldsymbol{\theta}_t))\|^2] \\
    &\quad + \cE[\|f_{\omega_\ell}^{(\ell)} \circ f_{\omega_{\ell-1}}^{(\ell-1)} \circ \cdots \circ f_{\omega_2}^{(2)}(F^{(1)}(\boldsymbol{\theta}_t)) - f_{\omega_\ell}^{(\ell)} \circ f_{\omega_{\ell-1}}^{(\ell-1)} \circ \cdots \circ f_{\omega_3}^{(3)}(F^{(2)}(\boldsymbol{\theta}_t))\|^2] + \cdots \\
    &\quad + \cE[\|f_{\omega_\ell}^{(\ell)}(F_t^{(\ell-1)}) - F_t^{(\ell)}\|^2] \Big) \\
    &\leq \ell \Big(\sprod_{i=2}^{\ell} \rho_i^2 \cE[\|f_{\omega_1}^{(1)}(\boldsymbol{\theta}_t) - F^{(1)}(\boldsymbol{\theta}_t)\|^2] + \sprod_{i=3}^{\ell} \rho_i^2 \cE[\|f_{\omega_2}^{(2)}(F^{(1)}(\boldsymbol{\theta}_t)) - F^{(2)}(\boldsymbol{\theta}_t)\|^2] \\
    &\qquad + \cdots + \cE[\|f_{\omega_\ell}^{(\ell)}(F^{(\ell-1)}(\boldsymbol{\theta}_t)) - F^{(\ell)}(\boldsymbol{\theta}_t)\|^2] \Big) \\
    &= \ell \sum_{i=1}^\ell \left(\sprod_{j=i+1}^\ell \rho_j^2 \cE[\|f_{\omega_j}^{(j)}(F^{(j-1)}(\boldsymbol{\theta}_t)) - F^{(j)}(\boldsymbol{\theta}_t)\|^2] \right).    
\end{aligned}
\end{equation}

Let $\mathbb{V}_\ell:= \cE[\|f_{\omega_\ell}^{(\ell)}(F^{(\ell-1)}(\boldsymbol{\theta}_t)) - F^{(\ell)}(\boldsymbol{\theta}_t)\|^2]$ as the per layer embedding approximation variance. Combining Eq.~\ref{eq:decouple_bias_1} and Eq.~\ref{eq:decouple_bias_2}, we obtain the upper bound on the bias of stochastic gradient $\cE[\|\mathbf{g}-\tilde{\mathbf{g}}\|^2]$ as a linear combination of per layer embedding approximation variance:
\begin{equation*}
    \cE[\|\mathbf{g}-\tilde{\mathbf{g}}\|^2] \leq L\cdot\sum_{\ell=2}^L \left(\sprod_{i=1}^{\ell-1} \rho^2_i \right)\left(\sprod_{i=\ell+1}^L \rho^2_i \right) G^2_\ell \cdot (\ell-1) \sum_{i=1}^{\ell-1} \left(\sprod_{j=i+1}^{\ell-1} \rho_j^2 \mathbb{V}_j \right) \leq L\cdot\sum_{\ell=2}^L \left(\sprod_{i=1}^{\ell-1} \rho^2_i \right)\left(\sprod_{i=\ell+1}^L \rho^2_i \right) G^2_\ell \cdot \ell \sum_{i=1}^{\ell} \left(\sprod_{j=i+1}^{\ell} \rho_j^2 \mathbb{V}_j \right). 
\end{equation*}
\label{section:bias_decouple_proof}
\section{Embedding Approximation Variance Analysis}\label{section:variance_analysis}
In this section, we analyze the variance of the approximation embedding for the sampled nodes at $\ell$th layer.

\begin{lemma} [Variance of \texttt{MVS-GNN}] \label{lemma:vars_mvs_gcn}
 We assume that for each node, \texttt{MVS-GNN} randomly sample $N_\ell$ nodes at $\ell$th layer to estimate the node embedding, then we have $\mathbb{V}_\ell \leq D \beta_\ell^2 \Delta \gamma^2_\ell$, where $D$ is the average node degree, $\Delta \gamma_\ell$ is the upper bound of $\|(\mathbf{H}_i^{(\ell-1)}-\bar{\mathbf{H}}_i^{(\ell-1)}) \mathbf{W}^{(\ell)}\|$, and $\beta_\ell$ is the upper bound of $\|\mathbf{L}_{i,*}\|$ for any $i\in\mathcal{V}$.
\end{lemma}
\begin{proof} [Proof of Lemma \ref{lemma:vars_mvs_gcn}]
By the update rule, we have
\begin{equation*}
    \begin{aligned}
        \mathbb{V}_\ell &= \cE[\|f_{\omega_\ell}^{(\ell)}(F^{(\ell-1)}(\boldsymbol{\theta}_t)) - F^{(\ell)}(\boldsymbol{\theta}_t)\|^2] \\
        &= \frac{1}{N_\ell} \sum_{i\in\mathcal{V}_\ell} \cE[\|\sum_{j\in\mathcal{V}_{\ell-1}} \tilde{L}_{i,j}^{(\ell)} \mathbf{H}_j^{(\ell-1)}\mathbf{W}^{(\ell)} + \sum_{j\in \mathcal{V}\backslash\mathcal{V}_{\ell-1}} L_{i,j} \bar{\mathbf{H}}_j^{(\ell-1)} \mathbf{W}^{(\ell)} - \sum_{j\in\mathcal{V}} L_{i,j} \mathbf{H}_j^{(\ell-1)} \mathbf{W}^{(\ell)} \|^2] \\
        &= \frac{1}{N_\ell} \sum_{i\in\mathcal{V}_\ell} \cE[\|\sum_{j\in\mathcal{V}_{\ell-1}} \tilde{L}_{i,j}^{(\ell)} \mathbf{H}_j^{(\ell-1)} \mathbf{W}^{(\ell)} + \sum_{j\in \mathcal{V}} L_{i,j} \bar{\mathbf{H}}_j^{(\ell-1)} \mathbf{W}^{(\ell)} - \sum_{j\in \mathcal{V}_{\ell-1}} \tilde{L}_{i,j}^{(\ell)} \bar{\mathbf{H}}_j^{(\ell-1)} \mathbf{W}^{(\ell)} - \sum_{j\in\mathcal{V}} L_{i,j} \mathbf{H}_j^{(\ell-1)} \mathbf{W}^{(\ell)} \|^2] \\
    \end{aligned}
\end{equation*}
Since \texttt{MVS-GNN} performs subgraph sampling, only the node in the mini-batch are guaranteed to be sampled in the inner layers. Therefore, the embedding approximation variance of \texttt{MVS-GNN} is similar to \texttt{VRGCN} with neighbor sampling size $s=1$. Denoting $\Delta \mathbf{H} = \mathbf{H}-\bar{\mathbf{H}}$, we have
\begin{equation*}
    \begin{aligned}
        \mathbb{V}_\ell &\leq \frac{1}{N_\ell} \sum_{i\in\mathcal{V}_\ell} \cE[\|\sum_{j\in\mathcal{V}_{\ell-1}} \tilde{\mathbf{L}}_{i,j}^{(\ell)} \Delta \mathbf{H}_j^{(\ell-1)} \mathbf{W}^{(\ell)} - \sum_{j\in\mathcal{V}} L_{i,j} \Delta \mathbf{H}_j^{(\ell-1)} \mathbf{W}^{(\ell)}\|^2] \\
        &= \frac{1}{N_\ell} \sum_{i\in\mathcal{V}_\ell} \left( D \sum_{j\in\mathcal{V}} \| \tilde{L}_{i,j}^{(\ell)} \Delta \mathbf{H}_j^{(\ell-1)} \mathbf{W}^{(\ell)}\|^2 -  \|\mathbf{L}_{i,*} \Delta \mathbf{H}^{(\ell-1)} \mathbf{W}^{(\ell)}\|^2\right) \\
        &\leq \frac{1}{N_\ell} \sum_{i\in\mathcal{V}_\ell} D \sum_{j\in\mathcal{V}} \| \tilde{L}_{i,j}^{(\ell)}\|^2 \|\Delta \mathbf{H}_j^{(\ell-1)} \mathbf{W}^{(\ell)}\|^2 \leq D  \beta^2_\ell \Delta\gamma^2_\ell
    \end{aligned}
\end{equation*}
\end{proof}

\section{Proof of Lemma~\ref{lemma:us_vs_is}}\label{section:us_vs_is}
\begin{figure}[h]
    \centering
    \includegraphics[width=0.6\textwidth]{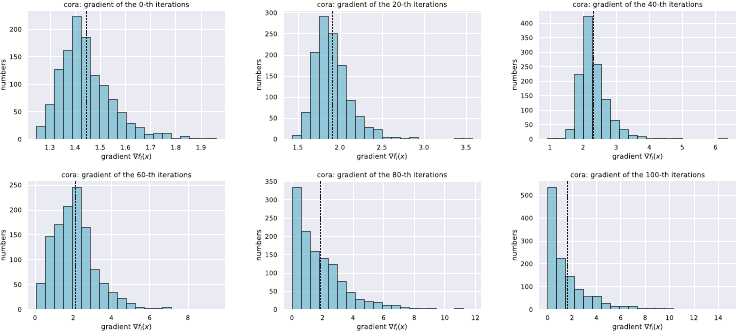}
    \vspace{-0.3cm}
    \caption{The per sample gradient distribution during training. }
    \label{fig:cora_grad}
\end{figure}

\begin{proof}
According to the definition of $\mathbb{G}(\boldsymbol{p}_{us})$ and $\mathbb{G}(\boldsymbol{p}_{is})$, we have
\begin{equation*}
    \begin{aligned}
        \mathbb{G}(\boldsymbol{p}_{us})-\mathbb{G}(\boldsymbol{p}_{is}) &= \frac{1}{N^2} \sum_{i=1}^N \frac{\bar{g}_i^2 N}{B} - \frac{1}{N^2} \sum_{i=1}^N \frac{\sum_{j=1}^N \bar{g}_j}{B\bar{g}_i} \bar{g}_i^2 \\
        &= \frac{1}{B} \sum_{i=1}^N \frac{\bar{g}_i^2}{N} - \frac{1}{B} \left(\sum_{i=1}^N \frac{\bar{g}_i}{N}\right)^2 \\
        &= \frac{(\sum_{i=1}^N \bar{g}_i)^2}{B N^3}\sum_{i=1}^N \left( N^2 \frac{\bar{g}_i^2}{(\sum_{j=1}^N \bar{g}_j)^2}-1\right) \\
        &= \frac{(\sum_{i=1}^N \bar{g}_i)^2}{B^3 N}\sum_{i=1}^N \left( \frac{B^2\bar{g}_i^2}{(\sum_{j=1}^N \bar{g}_j)^2}-\frac{B^2}{N^2}\right) \\
    \end{aligned}
\end{equation*}
Using the fact that $\sum_{i=1}^N 1/N = 1$, we complete the derivation.
\begin{equation*}
    \begin{aligned}
        \mathbb{G}(\boldsymbol{p}_{us})-\mathbb{G}(\boldsymbol{p}_{is}) 
        &= \frac{(\sum_{i=1}^N \bar{g}_i)^2}{B^3 N}\sum_{i=1}^N \left( \frac{B\bar{g}_i}{\sum_{j=1}^N \bar{g}_j}-\frac{B}{N}\right)^2 = \frac{(\sum_{i=1}^N \bar{g}_i)^2}{B^3 N} \|\boldsymbol{p}_{is}-\boldsymbol{p}_{us}\|_2^2.
    \end{aligned}
\end{equation*}
\end{proof}\label{section:variance_analysis_proof}
\noindent\textbf{Evaluation on gradient distribution.~}
To further illustrate the importance of minimal variance sampling, we show the distribution of per sampler gradient during training on \texttt{Cora} dataset in Figure~\ref{fig:cora_grad}, where the dash line stands for the full-batch gradient. We observe that certain stochastic gradients have more impact on the full-batch gradient than others, which motivates us to further reduce the variance of mini-bath  by sampling nodes with (approximately) large gradients more frequently.

\label{section:grad_dists}

\end{document}